\documentclass[10pt,twocolumn,letterpaper]{article}
\usepackage{xpatch}
\xpretocmd{\part}{\setcounter{section}{0}}{}{}
\usepackage{cvpr}
\usepackage{times}
\usepackage{textcomp}
\usepackage{epsfig}
\usepackage{graphicx}
\usepackage{amsmath}
\usepackage{amssymb}
\usepackage{hhline}
\usepackage{enumitem}
\usepackage{multirow}
\usepackage[normalem]{ulem}
\newenvironment{packed_enum}{
\begin{itemize}[leftmargin=*]
  \setlength{\itemsep}{0.6pt}
  \setlength{\parskip}{0pt}
  \setlength{\parsep}{0pt}
}{\end{itemize}}
\usepackage{lipsum} 
\usepackage{tabularx,colortbl}
\usepackage{float}
\usepackage{subcaption}
\usepackage[symbol]{footmisc}
\usepackage[figurename=Fig.]{caption}
\usepackage{makecell}
\floatstyle{plaintop}
\restylefloat{table}

\usepackage{soul} 
\newcolumntype{?}{!{\vrule width 1pt}}
\newcolumntype{P}[1]{>{\centering\arraybackslash}p{#1}}
\newcolumntype{C}[1]{>{\centering\arraybackslash}c{#1}}
\usepackage[pagebackref=true,breaklinks=true,letterpaper=true,colorlinks,bookmarks=false]{hyperref}

\newcommand\blfootnote[1]{%
  \begingroup
  \renewcommand\thefootnote{}\footnote{#1}%
  \addtocounter{footnote}{-1}%
  \endgroup
}

\cvprfinalcopy 

\def\cvprPaperID{6584} 

\ifcvprfinal\fi

\begin{document}

\renewcommand{\thefootnote}{\fnsymbol{footnote}}
\title{From Patches to Pictures (PaQ-2-PiQ): \\ Mapping the Perceptual Space of Picture Quality}



\author{
	Zhenqiang Ying\textsuperscript{1\footnotemark[1]}, 
	Haoran Niu\textsuperscript{1\footnotemark[1]}, 
	Praful Gupta\textsuperscript{1}, 
	Dhruv Mahajan\textsuperscript{2}, 
	Deepti Ghadiyaram\textsuperscript{2\footnotemark[2]}, 
	Alan Bovik\textsuperscript{1\footnotemark[2]}\\
	\textsuperscript{1}University of Texas at Austin, \textsuperscript{2}Facebook AI\\
	{\tt\small \{zqying, haoranniu, praful\_gupta\}@utexas.edu, \{dhruvm, deeptigp\}@fb.com, bovik@ece.utexas.edu}
}


\maketitle
\begin{abstract}
Blind or no-reference (NR) perceptual picture quality prediction is a difficult, unsolved problem of great consequence to the social and streaming media industries that impacts billions of viewers daily. Unfortunately, popular NR prediction models perform poorly on real-world distorted pictures. To advance progress on this problem, we introduce the largest (by far) subjective picture quality database, containing about $40000$ real-world distorted pictures and $120000$ patches, on which we collected about $4M$ human judgments of picture quality. Using these picture and patch quality labels, we built deep region-based architectures that learn to produce state-of-the-art global picture quality predictions as well as useful local picture quality maps. Our innovations include picture quality prediction architectures that produce global-to-local inferences as well as local-to-global inferences (via feedback). 
\end{abstract}


\blfootnote{\textsuperscript{$*\dagger$}Equal contribution}
\section{Introduction}
Digital pictures, often of questionable quality, have become ubiquitous. Several hundred billion photos are uploaded and shared annually on social media sites like Facebook, Instagram, and Tumblr. Streaming services like Netflix, Amazon Prime Video, and YouTube account for $60\%$ of all downstream internet traffic \cite{vidReport}. Being able to understand and predict the perceptual quality of digital pictures, given resource constraints and increasing display sizes, is a high-stakes problem. 

\begin{figure}[t]
\begin{center}
    $\begin{array}{ccc}
    \includegraphics[width=0.35\linewidth, height = 3.0cm]{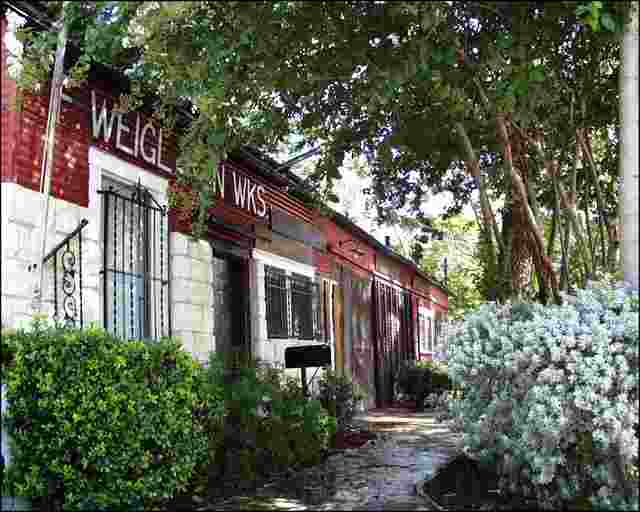} &
    \hspace{-0.8em}
    \includegraphics[width=0.25\linewidth, height = 3.0cm]{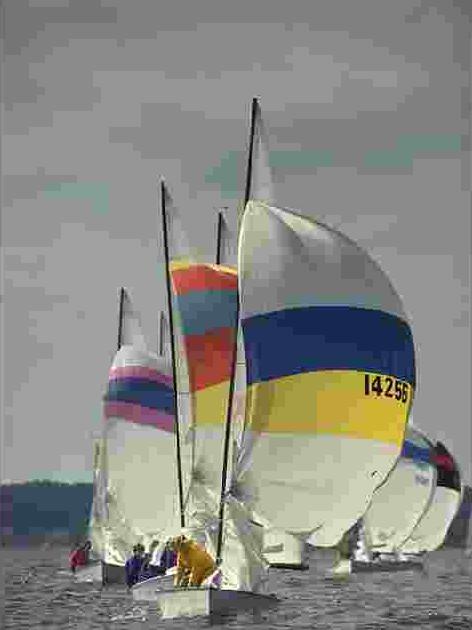} &
    \hspace{-0.8em}
    \includegraphics[width=0.35\linewidth, height = 3.0cm]{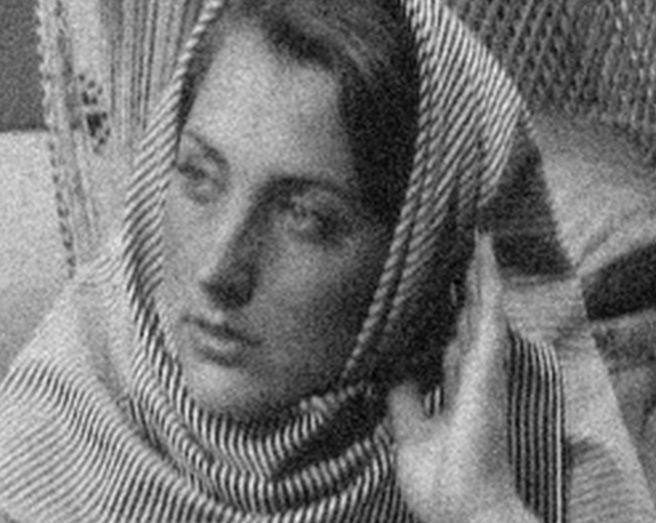} \\
    \scriptsize(a) & (b) & (c) \\
    \end{array}$  
    \vspace{-0.3cm}
    \captionof{figure}{\footnotesize{\textbf{Challenges in distortion perception:} Quality of a (distorted) image as perceived by human observers is \textit{perceptual quality}. Distortion perception is highly content-dependent. Pictures (a) and (b) were JPEG compressed using identical encode parameters, but present very different degrees of perceptual distortion. The spatially uniform noise in (c) varies in visibility over the picture content, because of contrast masking~\cite{autoBovik}.}}
    \label{fig:teaser}
\end{center}
\end{figure}

It is a common misconception that if two pictures are impaired by the same amount of a distortion (e.g., blur), they will have similar perceived qualities. However, this is far from true because of the way the vision system processes picture impairments. For example, Figs.~\ref{fig:teaser}(a) and (b) have identical amounts of JPEG compression applied, but Fig.~\ref{fig:teaser}(a) appears relatively unimpaired perceptually, while Fig.~\ref{fig:teaser}(b) is unacceptable. On the other hand, Fig.~\ref{fig:teaser}(c) has had spatially uniform white noise applied to it, but its perceived distortion severity varies across the picture. The complex interplay between picture content and distortions (largely determined by masking phenomena~\cite{autoBovik}), and the way distortion artifacts are visually processed, play an important role in how visible or annoying visual distortions may present themselves. Moreover, perceived quality correlates poorly with simple quantities like resolution and bit rate~\cite{mseLoveLeave}. Generally, predicting perceptual picture quality is a hard, long-standing research problem~\cite{mannos, autoBovik, mseLoveLeave, ssim, vif}, despite its deceptive simplicity (we sense distortion easily with little, if any, thought).

\begin{figure}[t]
\centering
\vspace{-0.2in}
$\begin{array}{cc}
\includegraphics[width=3.6cm, height = 3.6cm]{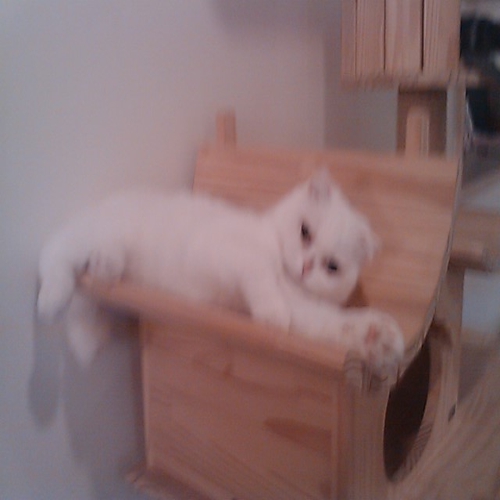} &
    \includegraphics[width=0.36\linewidth, height = 3.6cm]{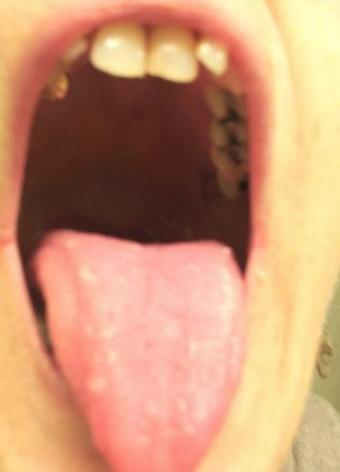} \\
    \scriptsize(a) & (b)\\
    \vspace{-2em}
\end{array}$
\caption{\footnotesize{\textbf{Aesthetics vs. perceptual quality} (a) is blurrier than (b), but likely  more aesthetically pleasing to most viewers.}}
\label{fig:aesthetics}
\vspace{-0.23in}
\end{figure}

It is important to distinguish between the concepts of \textit{picture quality} \cite{autoBovik} and \textit{picture aesthetics} \cite{ava}. Picture quality is specific to perceptual distortion, while aesthetics also relates to aspects like subject placement, mood, artistic value, and so on. For instance, Fig.~\ref{fig:aesthetics}(a) is noticeably blurred and of lower perceptual quality than Fig.~\ref{fig:aesthetics}(b), which is less distorted. Yet, Fig.~\ref{fig:aesthetics}(a) is more aesthetically pleasing than the unsettling Fig.~\ref{fig:aesthetics}(b). While distortion can detract from aesthetics, it can also contribute to it, as when intentionally adding film grain \cite{filmGrain} or blur (bokeh) \cite{bokeh} to achieve photographic effects. While both concepts are important, picture quality prediction is a critical, high-impact problem affecting several high-volume industries, and is the focus of this work. Robust picture quality predictors can significantly improve the visual experiences of social media, streaming TV and home cinema, video surveillance, medical visualization, scientific imaging, and more.

In many such applications, it is greatly desired to be able to assess picture quality at the point of ingestion, to better guide decisions regarding retention, inspection, culling, and all further processing and display steps. Unfortunately, measuring picture quality without a pristine \textit{reference} picture is very hard. This is the case at the output of any camera, and at the point of content ingestion by any social media platform that accepts user-generated content (UGC). \textit{No-reference} (NR) or blind picture quality prediction is largely unsolved, though popular models exist \cite{mittal2012no,niqe, deeptiBag,cornia,hosa,nferm, qac}. While these are often predicated on solid principles of visual neuroscience, they are also simple and computationally shallow, and fall short when tested on recent databases containing difficult, complex mixtures of real-world picture distortions~\cite{clive, koniq}. Solving this problem could affect the way billions of pictures uploaded daily are culled, processed, compressed, and displayed.

Towards advancing progress on this high-impact unsolved problem, we make several new contributions.
\vspace{-0.1in}
\begin{packed_enum}
\item \textbf{We built the largest picture quality database in existence}. We sampled hundreds of thousands of open source digital pictures to match the feature distributions of the largest use-case: pictures shared on social media. The final collection includes about $40,000$ real-world, unprocessed (by us) pictures of diverse sizes, contents, and distortions, and about $120,000$ cropped image patches of various scales and aspect ratios (Sec.~\ref{sec:picture_sampling},~\ref{sec:patch_cropping}).
\item \textbf{We conducted the largest subjective picture quality study to date.} We used Amazon Mechanical Turk to collect about $4M$ human perceptual quality judgments from almost $8,000$ subjects on the collected content, about four times more than any prior image quality study (Sec.~\ref{sec:crowdsourcing}).
\item \textbf{We collected both picture and patch quality labels to relate local and global picture quality}. The new database includes about $1M$ human picture quality judgments and $3M$ human quality labels on patches \textit{drawn from the same pictures}. 
Local picture quality is deeply related to global quality, although this relationship is not well understood \cite{moorthyPooling}, \cite{vpooling}. This data will help us to learn these relationships and to better model global picture quality. 
\item \textbf{We created a series of state-of-the-art deep blind picture quality predictors}, that builds on existing deep neural network architectures. Using a modified ResNet~\cite{resNet} as a baseline, we (a) use patch and picture quality labels to train a region proposal network~\cite{fastRCNN},~\cite{fasterRCNN} to predict both global picture quality and local patch quality. This model is able to produce better global picture quality predictions by learning relationships between global and local picture quality (Sec.~\ref{sec:p2p_model}). We then further modify this model to (b) predict spatial maps of picture quality, useful for localizing picture distortions (Sec.~\ref{sec:qualityMaps}). Finally, we (c) innovate a local-to-global feedback architecture that produces further improved whole picture quality predictions using local patch predictions (Sec.~\ref{subsec:patchAugQuality}). This series of models obtains state-of-the art picture quality performance on the new database, and transfer well --  \emph{without finetuning} -- on smaller ``in-the-wild” databases such as LIVE Challenge (CLIVE) \cite{clive} and KonIQ-10K \cite{koniq} (Sec.~\ref{sec:cross_data}).

\end{packed_enum}
\section{Background}\label{sec:background}
\begin{table*}[t]
\captionsetup{font=scriptsize}
\vspace{-0.1in}
\scriptsize
\centering
\begin{tabular}{|c|c|c|c|c|c|c|c|c|}
\hline
Database & \thead{\scriptsize{\# Unique} \\ \scriptsize{contents}} & \# Distortions & \thead{\scriptsize{\# Picture} \scriptsize{contents}} & \thead{\scriptsize{\# Patch} \scriptsize{contents}} & Distortion type & \thead{\scriptsize{Subjective study} \\ \scriptsize{framework}} & \# Annotators &
\# Annotations  \\
\hline
LIVE IQA (2003)~\cite{hamidLiveDB} & 29 & 5 & 780 & 0 & single, synthetic & in-lab & & \\

TID-2008~\cite{ponomarenko2009tid2008} & 25 & 17 & 1700 & 0 & single, synthetic & in-lab & & \\

TID-2013~\cite{ponomarenko2009tid2008} & 25 & 24 & 3000 & 0 & single, synthetic & in-lab & &\\
\hline
CLIVE (2016)~\cite{clive}  & 1200 & - & 1200 & 0 & in-the-wild & crowdsourced & $8000$ &$350$K \\

KonIQ (2018)~\cite{koniq} & $10$K & - & $10$K & 0 & in-the-wild & crowdsourced & $1400$ &$1.2$M  \\
\hline
\textbf{Proposed database} &  $39,810$ & - & $39,810$ & $119,430$ &  in-the-wild &  crowdsourced & $7865$ & $3,931,710$ \\
\hline
\end{tabular}
\caption{\scriptsize{\textbf{Summary of popular IQA datasets.} In the legacy datasets, pictures were synthetically distorted with different types of single distortions.
``In-the-wild'' databases contain pictures impaired by complex mixtures of highly diverse distortions, each as unique as the pictures they afflict.}} 
\vspace{-1.5em}
\label{tbl:datasets}
\end{table*}
\noindent\textbf{Image Quality Datasets:} Most picture quality models have been designed and evaluated on three ``legacy" databases: LIVE IQA~\cite{hamidLiveDB}, TID-2008~\cite{ponomarenko2009tid2008}, and TID-2013~\cite{tid2013}. These datasets contain small numbers of unique, pristine images ($\sim30$) synthetically distorted by diverse types and amounts of single distortions (JPEG, Gaussian blur, etc.). They contain limited content and distortion diversity, and do not capture complex mixtures of distortions that often occur in real-world images. Recently, ``in-the-wild'' datasets such as CLIVE~\cite{clive} and KonIQ-$10$K~\cite{koniq}, have been introduced to attempt to address these shortcomings (Table \ref{tbl:datasets}).

\noindent\textbf{Full-Reference models:} Many \textit{full-reference} (FR) perceptual picture quality predictors, which make comparisons against high-quality \textit{reference} pictures, are available~\cite{ssim, vif},~\cite{msssim, mad, fsim, vsnr, gmsd, haariqa, vsi}. Although some FR algorithms (e.g. SSIM~\cite{ssim},~\cite{emmyAward}, VIF~\cite{vif},~\cite{vmafEncodeNetflixBlog, shotEncodeNetflixBlog}) have achieved remarkable commercial success (e.g. for monitoring streaming content), they are limited by their requirement of pristine reference pictures.

\noindent\textbf{Current NR models aren’t general enough:} 
\textit{No-reference} or blind algorithms predict picture content without the benefit of a reference signal. Popular blind picture quality algorithms usually measure distortion-induced deviations from perceptually relevant, highly regular bandpass models of picture statistics \cite{autoBovik}, \cite{field87, ruderman1994, simoncelli2001natural, textureBovik}. Examples include BRISQUE \cite{mittal2012no}, NIQE \cite{niqe}, CORNIA \cite{cornia}, FRIQUEE \cite{deeptiBag}, which use ``handcrafted'' statistical features to drive shallow learners (SVM, etc.). These models produce accurate quality predictions on legacy datasets having single, synthetic distortions~\cite{hamidLiveDB, ponomarenko2009tid2008, tid2013, csiq}, but struggle on recent “in-the-wild”~\cite{clive, koniq} databases.

Several deep NR models \cite{ghadiyaram2014blind, deepConvIQA, bosseDeepIQA, fullyDeepIQA, nima} have also been created that yield state-of-the-art performance on legacy synthetic distortion databases \cite{hamidLiveDB, ponomarenko2009tid2008, tid2013, csiq}, e.g., by pretraining deep nets \cite{simonyan2014deep, rankIQA, kedeMaIQA} on ImageNet \cite{imageNet}, then fine tuning, or by training on proxy labels generated by an FR model \cite{fullyDeepIQA}. However, most deep models also struggle on CLIVE~\cite{clive}, because it is too difficult, yet too small to sufficiently span the perceptual space of picture quality to allow very deep models to map it. The authors of \cite{Bianco2018}, the code of which is not made available, reported high results, but we have been unable to reproduce their numbers, even with more efficient networks. The authors of \cite{Varga2018DeeprnAC} use a pre-trained ResNet-101 and report high performance on~\cite{clive, koniq}, but later disclosed \cite{saupePage} that they are unable to reproduce their own results in \cite{Varga2018DeeprnAC}.
\section{Large-Scale Dataset and Human Study}\label{sec:dataset}

\begin{figure}[t]
\begin{center}$
\begin{array}{cccc}
\hspace{-0.09in}
\includegraphics[height= 0.22\linewidth, width=0.28\linewidth]{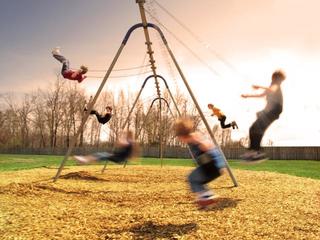} &
\hspace{-0.13in}
\includegraphics[height= 0.22\linewidth, width=0.15\linewidth]{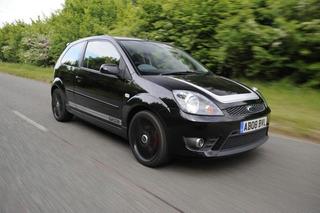} &
\hspace{-0.13in}
\includegraphics[height= 0.22\linewidth, width=0.28\linewidth]{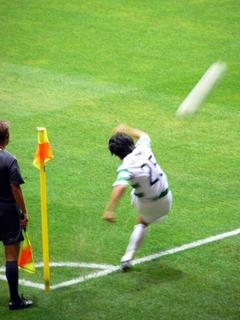} &
\hspace{-0.13in}
\includegraphics[height= 0.22\linewidth, width=0.28\linewidth]{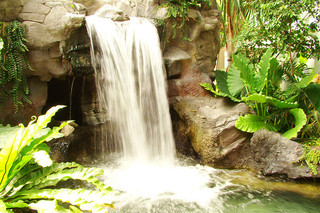} \\
\end{array}$ 

\vspace{-0.02in}

$\begin{array}{cccc}
\hspace{-0.09in}
\includegraphics[height= 0.22\linewidth, width=0.15\linewidth]{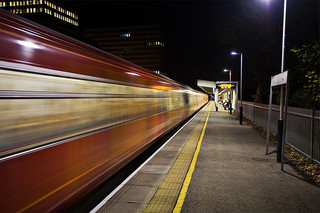} &
\hspace{-0.13in}
\includegraphics[height= 0.22\linewidth, width=0.28\linewidth]{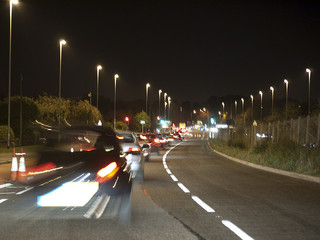} &
\hspace{-0.13in}
\includegraphics[height= 0.22\linewidth, width=0.28\linewidth]{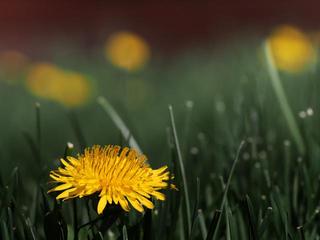} &
\hspace{-0.13in}
\includegraphics[height= 0.22\linewidth, width=0.28\linewidth]{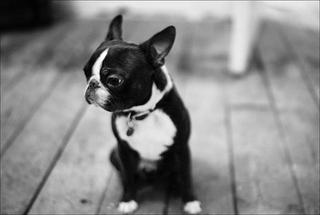} \\
\end{array}$

\vspace{-0.02in}

$\begin{array}{cccc}
\hspace{-0.09in}
\includegraphics[height= 0.22\linewidth, width=0.28\linewidth]{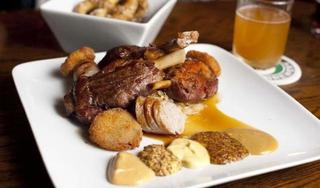} &
\hspace{-0.13in}
\includegraphics[height= 0.22\linewidth, width=0.28\linewidth]{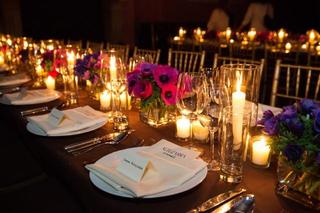} &
\hspace{-0.13in}
\includegraphics[height= 0.22\linewidth, width=0.15\linewidth]{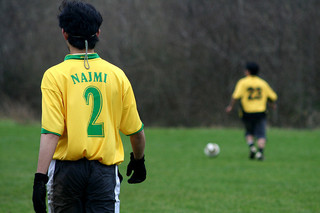} &
\hspace{-0.13in}
\includegraphics[height= 0.22\linewidth, width=0.28\linewidth]{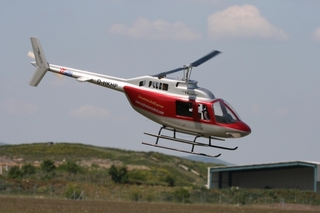} \\
\end{array}$

\vspace{-0.02in}

$\begin{array}{cccc}
\hspace{-0.09in}
\includegraphics[height= 0.22\linewidth, width=0.15\linewidth]{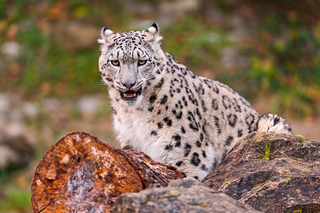} &
\hspace{-0.13in}
\includegraphics[height= 0.22\linewidth, width=0.28\linewidth]{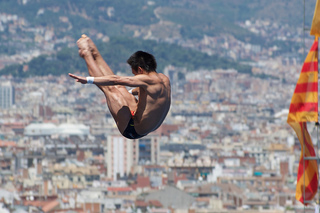} &
\hspace{-0.13in}
\includegraphics[height= 0.22\linewidth, width=0.28\linewidth]{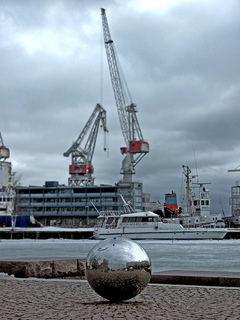} &
\hspace{-0.13in}
\includegraphics[height= 0.22\linewidth, width=0.28\linewidth]{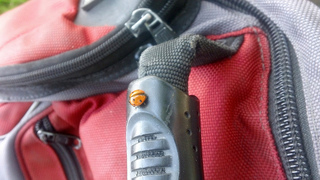} \\
\end{array}$
\vspace{-0.15in}
\caption{\scriptsize{\textbf{Exemplar pictures from the new database}, each resized to fit. Actual pictures are of highly diverse sizes and shapes.}}
\vspace{-2em}
\label{fig:exemplarFLIVE}
\end{center}
\end{figure}
Next we explain the details of the new picture quality dataset we constructed, and the crowd-sourced subjective quality study we conducted on it. The database has about $40,000$ pictures and $120,000$ patches, on which we collected $4$M human judgments from nearly $8,000$ unique subjects (after subject rejection). It is significantly larger than commonly used ``legacy” databases~\cite{hamidLiveDB, ponomarenko2009tid2008, tid2013, csiq} and more recent ``in-the-wild'' crowd-sourced datasets~\cite{clive, koniq}. 
\subsection{UGC-like picture sampling} \label{sec:picture_sampling}
Data collection began by sampling about $40$K highly diverse contents of diverse sizes and aspect ratios from hundreds of thousands of pictures drawn from public databases, including AVA \cite{ava}, VOC \cite{voc}, EMOTIC \cite{emotic}, and CERTH Blur \cite{certh}. Because we were interested in the role of local quality perception as it relates to global quality, we also cropped \underline{three} patches from each picture, yielding about $120$K patches. While internally debating the concept of ``representative,” we settled on a method of sampling a large image collection so that it would be substantially ``UGC-like.” We did this because billions of pictures are uploaded, shared, displayed, and viewed on social media, far more than anywhere else. 
\begin{figure}[h]
\vspace{-1em}
\begin{center}
\includegraphics[width=0.7\linewidth]{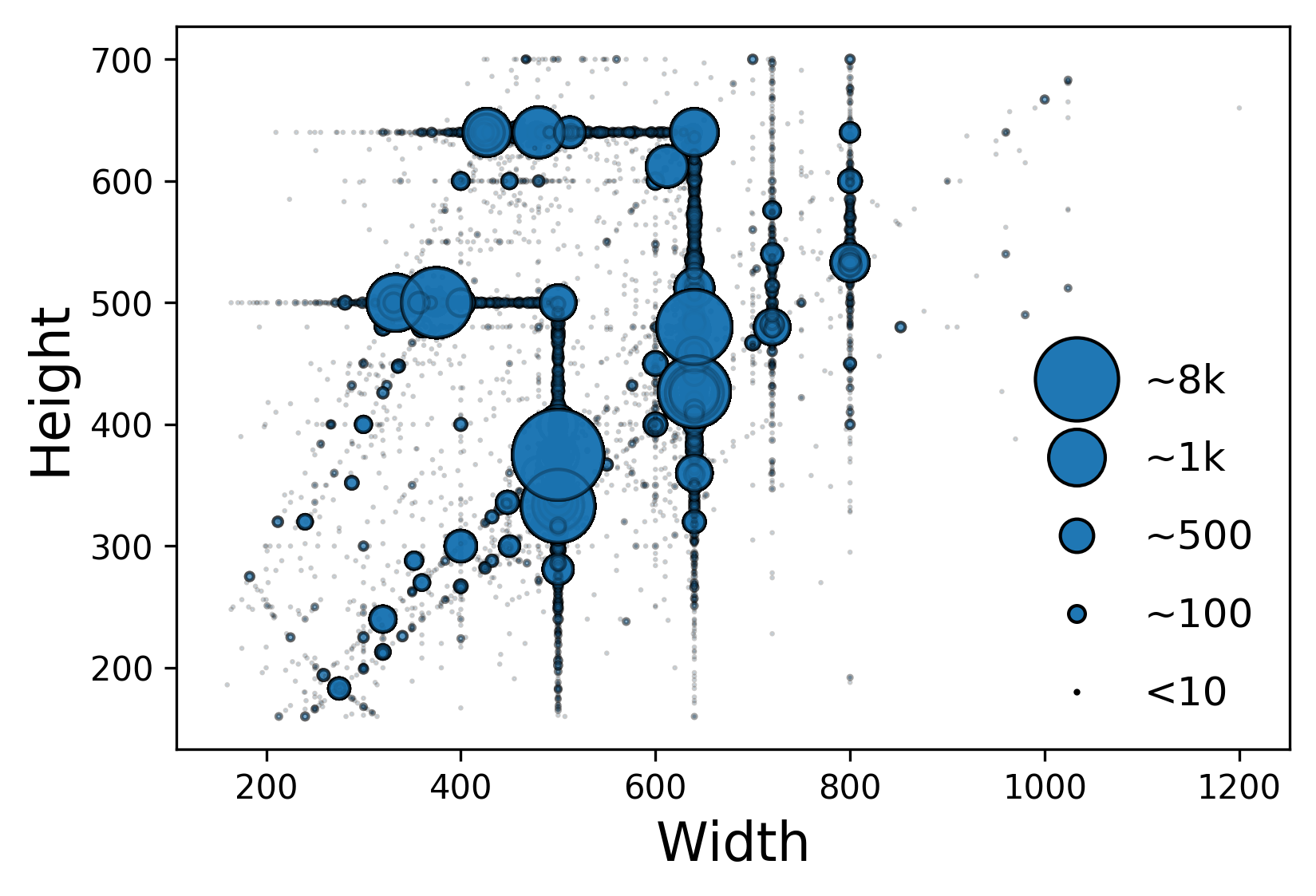}
\vspace{-1em}
\caption{\scriptsize{\textbf{Scatter plot of picture width versus picture height} with marker size indicating the number of pictures for a given dimension in the new database.}}
\vspace{-2em}
\label{fig:pixel_aspect}
\end{center}
\end{figure}

We sampled picture contents using a mixed integer programming method~\cite{mixedInteger} similar to~\cite{koniq}, to match a specific set of UGC feature histograms.
Our sampling strategy was different in several ways: firstly, unlike KonIQ~\cite{koniq}, no pictures were down sampled, since this intervention can substantially modify picture quality. 
Moreover, including pictures of diverse sizes better reflects actual practice. Second, instead of uniformly sampling feature values, we designed a picture collection whose feature histograms match those of $15$M randomly selected pictures from a social media website. This in turn resulted in a much more realistic and difficult database to predict features on, as we will describe later. Lastly, we did not use a pre-trained IQA algorithm to aid the picture sampling, as that could introduce \textit{algorithmic bias} into the data collection process. 

To sample and match feature histograms, we computed the following diverse, objective features on both our picture collection and the $15$M UGC pictures: 
\begin{packed_enum}
\vspace{-0.05in}
\item \textit{absolute brightness} $L = R + G + B$. 
\item \textit{colorfulness} using the popular model in \cite{measureColor}. 
\item \textit{RMS brightness contrast} \cite{peli}. 
\item \textit{Spatial Information(SI)}, the global standard deviation of Sobel gradients \cite{yu2013image}, a measure of complexity. 
\item \textit{pixel count}, a measure of picture size. 
\item number of \textit{detected faces} using~\cite{faceDetection}.
\vspace{-0.05in}
\end{packed_enum}
In the end, we arrived at about $40$K pictures. Fig. \ref{fig:exemplarFLIVE} shows $16$ randomly selected pictures and Fig.~\ref{fig:pixel_aspect} highlights the diverse sizes and aspect ratios of pictures in the new database.

\subsection{Patch cropping} \label{sec:patch_cropping}
\begin{figure}[h]
\vspace{-0.1in}
\begin{center}$
\begin{array}{cc}
\includegraphics[width=0.48\linewidth]{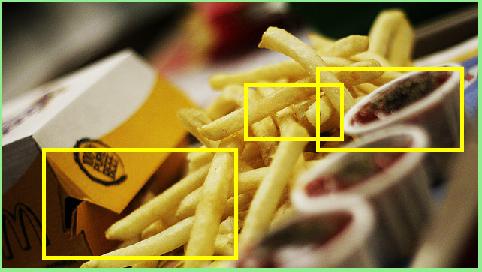} & \hspace{-0.7em} \includegraphics[width=0.48\linewidth]{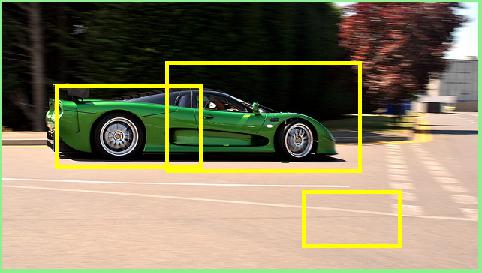} \\
\end{array}$ 
\vspace{-0.15in}
\caption{\scriptsize{Sample pictures and $3$ randomly positioned crops ($20\%$, $30\%$, $40\%$).}}
\vspace{-0.3in}
\label{fig:egImagesPatches}
\end{center}
\end{figure}

We applied the following criteria when randomly cropping out patches: (a) \textbf{aspect ratio:} patches have the same aspect ratios as the pictures they were drawn from. (b) \textbf{dimension:} the linear dimensions of the patches are $40\%$, $30\%$, and $20\%$ of the picture dimensions. (c) \textbf{location:} every patch is entirely contained within the picture, but no patch  overlaps the area of another patch cropped from the same image by more than $25\%$. Fig. \ref{fig:egImagesPatches} shows two exemplar pictures, and three patches obtained from each.

\begin{figure}[t]
\begin{center}
\includegraphics[width=0.8\linewidth, trim={0em 0em 1.5em 0em},clip]{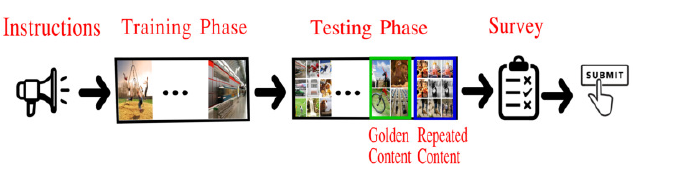}
\vspace{-2em}
\caption{\scriptsize{\textbf{AMT task:} Workflow experienced by crowd-sourced workers when rating either pictures or patches.}}
\vspace{-2em}
\label{fig:AMTdesign}
\end{center}
\end{figure}

\subsection{Crowdsourcing pipeline for subjective study} \label{sec:crowdsourcing}
Subjective picture quality ratings are true psychometric measurements on human subjects, requiring $10$-$20$ times as much time for scrutiny (per photo) as for example, object labelling \cite{imageNet}. We used the Amazon Mechanical Turk (AMT) crowdsourcing system, well-documented for this purpose \cite{clive, koniq, amt, cliveVideo}, to gather human picture quality labels. 

We divided the study into two separate tasks: picture quality evaluation and patch quality evaluation. Most subjects ($7141$ out of $7865$ workers) only participated in one of these, to avoid biases incurred by viewing both, even on different dates. 
Either way, the crowdsource workflow was the same, as depicted in Fig. \ref{fig:AMTdesign}. Each worker was given instructions, followed by a training phase, where they were shown several contents to learn the rating task. They then viewed and quality-rated $N$ contents to complete their human intelligent task (HIT), concluding with a survey regarding their experience. At first, we set $N = 60$, but as the study accelerated and we found the workers to be delivering consistent scores, we set $N = 210$. We found that the workers performed as well when viewing the increased number of pictures.
\subsection{Processing subjective scores}\label{subsec:processScores}
\noindent \textbf{Subject rejection:} We took the recommended steps~\cite{clive, cliveVideo} to ensure the quality of the collected human data. 
\vspace{-0.08in}
\begin{packed_enum}
\item We only accepted workers with \textbf{acceptance rates} $>75\%$. \item \textbf{Repeated images:} $5$ of the $N$ contents were repeated randomly per session to determine whether the subjects were giving consistent ratings.
\item \textbf{``Gold" images:} $5$ out of $N$ contents were ``gold” ones sampled from a collection of $15$ pictures and $76$ patches that were separately rated in a controlled lab study by $18$ reliable subjects. The ``gold" images are not part of the new database.
\end{packed_enum}
\vspace{-0.08in}
We accepted or rejected each rater’s scores within a HIT based on two factors: the difference of the repeated content scores compared with overall standard deviation, and whether more than $50\%$ of their scores were identical. Since we desired to capture many ratings, workers could participate in multiple HITs. Each content received at least $35$ quality ratings, with some receiving as many as $50$. 

The labels supplied by each subject were converted into normalized Z scores \cite{hamidLiveDB}, \cite{clive}, averaged (by content), then scaled to [0, 100] yielding \textbf{Mean Opinion Scores (MOS)}. 
The total number of human subjective labels collected after subject rejection was $3,931,710$ ($950,574$ on images, and $2,981,136$ on patches). 

\noindent\textbf{Inter-subject consistency:} A standard way to test the consistency of subjective data \cite{hamidLiveDB}, \cite{clive}, is to randomly divide subjects into two disjoint equal sets, compute two MOS on each picture (one from each group), then compute the Pearson linear correlation (LCC) between the MOS values of the two groups. When repeated over $25$ random splits, the average LCC between the two groups’ MOS was $\mathbf{0.48}$, indicating the difficulty of the quality prediction problem on this realistic picture dataset.
Fig. \ref{fig:patchCorrel} (left) shows a scatter plot of the two halves of human labels for one split, showing a linear relationship and fairly broad spread. We applied the same process to the patch scores, obtaining a higher LCC of $\mathbf{0.65}$. This is understandable: smaller patches contain less spatial diversity; hence they receive more consistent scores. We also found that nearly all the non-rejected subjects had a positive Spearman rank ordered correlation (SRCC) with the golden pictures, validating the data collection process.

\noindent\textbf{Relationships between picture and patch quality:} Fig. \ref{fig:patchCorrel} (right) is a scatter plot of the entire database of picture MOS against the MOS of the largest patches cropped from them. The linear correlation coefficient (LCC) between them is $0.43$, which is strong, given that each patch represents only $16\%$ of the picture area. The scatter plots of the picture MOS against that of the smaller ($30\%$ and $20\%$) patches are quite similar, with somewhat reduced LCC of $0.36$ and $0.28$, respectively (supplementary material).

An outcome of creating highly realistic ``in the wild” data is that it is much more difficult to train successful models on. Most pictures uploaded to social media are of reasonably good quality, largely owing to improved mobile cameras.
\begin{figure}[t]
\begin{center}
$\begin{array}{cc}
\includegraphics[height= 0.35\linewidth,width=0.46\linewidth]{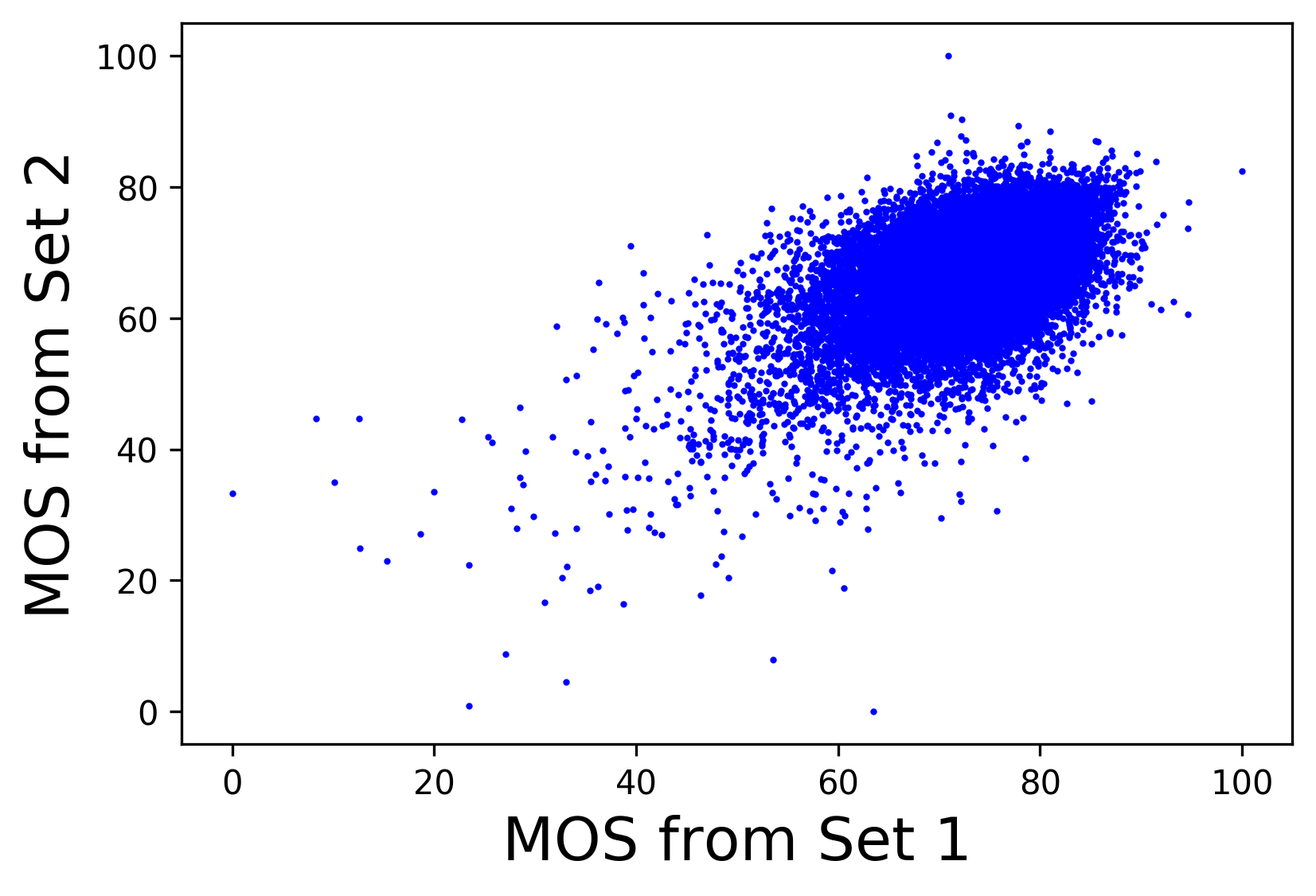} & 
\hspace{-1em}
\includegraphics[height= 0.35\linewidth,width=0.46\linewidth]{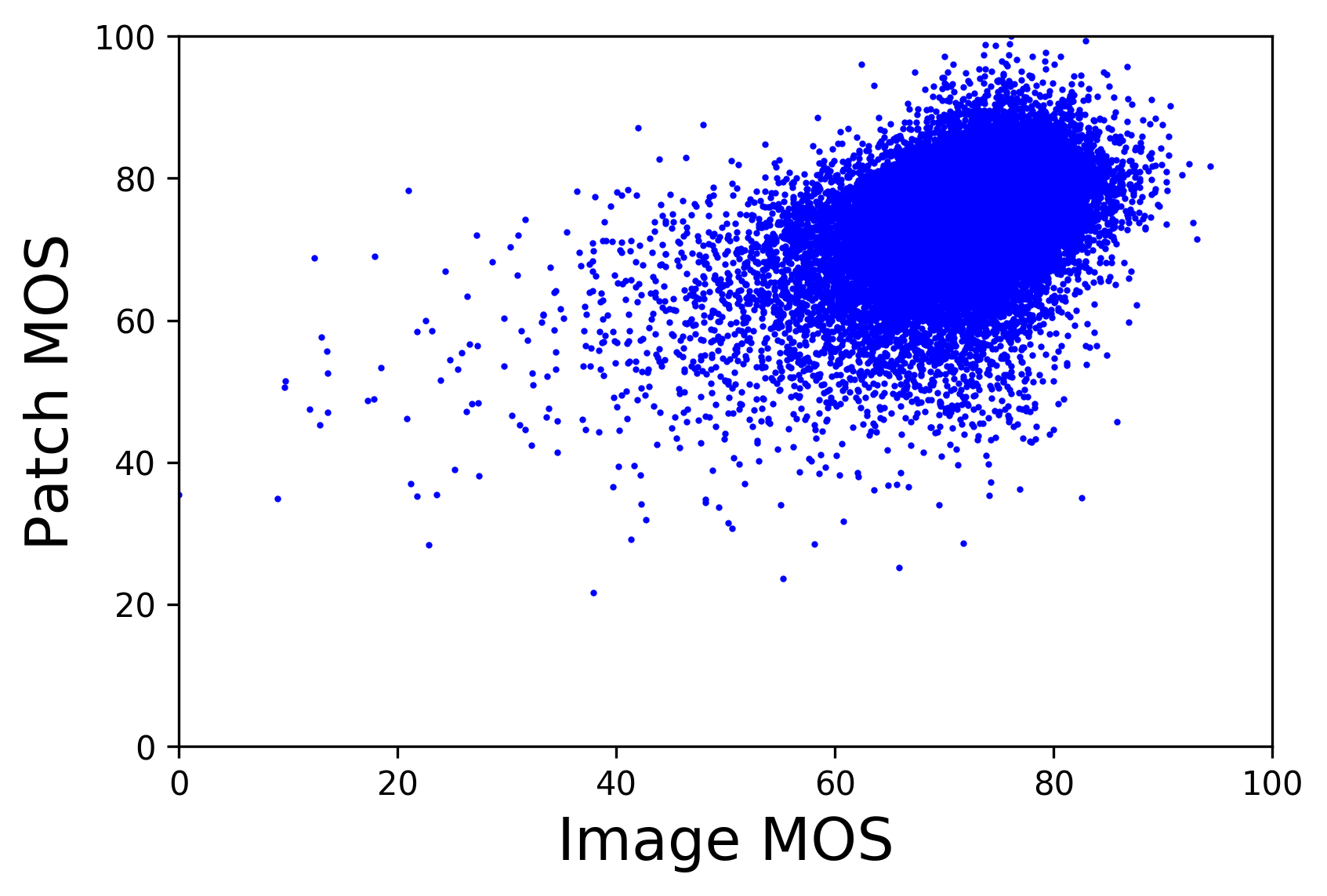} \\
\end{array}$ 
\vspace{-1em}
\caption{\scriptsize{\textbf{Scatter plots descriptive of the new subjective quality database}. Left: Inter-subject scatter plot of a random $50\%$ divisions of the human labels of all $40$K$+$ pictures into disjoint subject sets. Right: Scatter plot of picture MOS vs MOS of largest patch ($40\%$ of linear dimension) cropped from each same picture. 
}}
\vspace{-2em}
\label{fig:patchCorrel}
\end{center}
\end{figure}
Hence, the distribution of MOS in the new database is narrower and peakier as compared to those of the two previous ``in the wild” picture quality databases \cite{clive}, \cite{koniq}.
This is important, since it is desirable to be able to predict small changes in MOS, which can be significant regarding, for example, compression parameter selection \cite{compressedClive}. As we show in Sec.~\ref{sec:modeling}, the new database is very challenging, even for deep models.
\begin{figure}[h]
\begin{center}
\includegraphics[width=\linewidth]{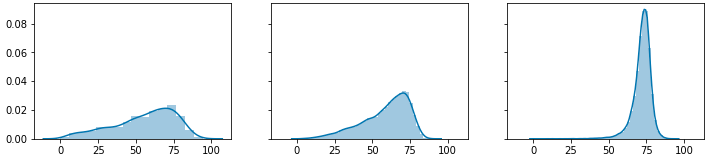}
\caption{\scriptsize{\textbf{MOS (Z-score) histograms of three ``in-the-wild” databases}. Left: CLIVE \cite{clive}. Middle: KoniIQ-$10$K \cite{koniq}. Right: The new database introduced here.}}
\vspace{-0.3in}
\label{fig:MOShists}
\end{center}
\end{figure}

\section{Learning Blind Picture Quality Predictors}\label{sec:modeling}
With the availability of the new dataset comprising pictures and patches associated with human labels (Sec.~\ref{sec:dataset}), we created a series of deep quality prediction models that exploit its unique characteristics. We conducted four picture quality learning experiments, evolving from a simple network into models of increasing sophistication and perceptual relevance which we describe next.
\subsection{A baseline picture-only model}\label{sec:baseline}
To start with, we created a simple model that only processes pictures and the associated human quality labels. We will refer to this hereafter as the Baseline Model. The basic network that we used is the well-documented pre-trained ResNet-$18$~\cite{resNet}, which we modified (described next) and fine-tuned to conduct the quality prediction task.

\noindent \textbf{Input image pre-processing:} Because picture quality prediction (whether by human or machine) is a psychometric prediction, it is crucial to not modify the pictures being fed into the network. While most visual recognition learners augment input images by cropping, resizing, flipping, etc., doing the same when training a perceptual quality predictor would be a psychometric error. Such input pre-processing would result in perceptual quality scores being associated with different pictures than they were recorded on. 

The new dataset contains thousands of unique combinations of picture sizes and aspect ratios (see Fig.~\ref{fig:pixel_aspect}). While this is a core strength of the dataset and reflects its realism, it also poses additional challenges when training deep networks. We attempted several ways of training the ResNet on raw multi-sized pictures, but the training and validation losses were not stable, because of the fixed sized pooling and fully connected layers.

In order to tackle this aspect, we white padded each training picture to size $640\times640$, centering the content in each instance. Pictures having one or both dimensions larger than $640$ were moved to the test set. This approach has the following advantages: (a) it allows supplying constant-sized pictures to the network, causing it to stably converge well, (b) it allows large batch sizes which improves training, (c) it agrees with the experiences of the picture raters, since AMT renders white borders around pictures that do not occupy the full webpage's width.

\noindent \textbf{Training setup:} We divided the picture dataset (and associated patches and scores) into training, validation and testing sets. Of the collected $39,810$ pictures (and $119,430$ patches), we used about $75\%$ for training ($30$K pictures, along with their $90$K patches), $19\%$ for validation ($7.7$K pictures, $23.1$K patches), and the remaining for testing ($1.8$K pictures, $5.4$K patches). When testing on the validation set, the pictures fed to the trained networks were also white bordered to size $640\times640$. As mentioned earlier, the test set is entirely composed of pictures having at least one linear dimension exceeding $640$. Being able to perform well on larger pictures of diverse aspect ratios was deemed as an additional challenge to the models. 

\noindent \textbf{Implementation Details:} We used the PyTorch implementation of ResNet-$18$~\cite{torchVision} pre-trained on ImageNet and retained only the CNN backbone during fine-tuning. To this, we added two pooling layers (adaptive average pooling and adaptive max pooling), followed by two fully-connected (\textit{FC}) layers, such that the final \textit{FC} layer outputs a single score. We used a batch size of $120$ and employed the MSE loss when regressing the single output quality score. We employed the Adam optimizer with $\beta_1=0.9$ and $\beta_2= 0.99$, a weight decay of $0.01$, and do a full fine-tuning for $10$ epochs. We followed a discriminative learning approach~\cite{howard2018universal}, using a lower learning rate of $3e^{-4}$, but a higher learning rate of $3e^{-3}$ for the head layers. These settings apply to all the models we describe in the following.

\noindent \textbf{Evaluation setup:} Although the baseline model was trained on whole pictures, we tested it on both pictures and patches. For comparison with popular shallow methods, we also trained and tested BRISQUE~\cite{mittal2012no} and the ``completely blind” NIQE~\cite{niqe}, which does not involve any training. We reimplemented two deep picture quality methods - NIMA
~\cite{nima} which uses a Mobilenet-v2~\cite{mobileNetV2} (except we replaced the output layer to regress a single quality score),
and CNNIQA~\cite{cnnIqa}, following the details provided by the authors. As is the common practice in the field of picture quality assessment, we report two metrics: (a) Spearman Rank Correlation Coefficient (\textbf{SRCC}) and (b) Linear Correlation Coefficient (\textbf{LCC}). 

\noindent \textbf{Results:} From Table \ref{tbl:onFlive}, the first thing to notice is the level of performance attained by popular shallow models (NIQE~\cite{niqe} and BRISQUE~\cite{mittal2012no}), which have the same feature sets. The unsupervised NIQE algorithm performed poorly, while BRISQUE did better, yet the reported correlations are far below desired levels. Despite being CNN-based, CNNIQA~\cite{cnnIqa} performed worse than BRISQUE~\cite{mittal2012no}. Our Baseline Model outperformed most methods and competed very well with NIMA~\cite{nima}. The other entries in the table (the ROIPool and Feedback Models) are described later. 
\begin{table}[t]
\captionsetup{font=scriptsize}
\setlength\extrarowheight{1.0pt}
\centering
\footnotesize
\vspace{0.8em}
\begin{tabular}{P{3.1cm}|P{0.85cm}|P{0.85cm}|P{0.85cm}|P{0.85cm}}
\hline
& \multicolumn{2}{c|}{\textbf{Validation Set}} & \multicolumn{2}{c}{\textbf{Testing Set}} \\
\hline
\textbf{Model} & \textbf{SRCC} & \textbf{LCC} & \textbf{SRCC} & \textbf{LCC} \\
\hline
NIQE \cite{niqe} & 0.094 & 0.131 & 0.211 & 0.288 \\
BRISQUE \cite{mittal2012no} & 0.303 & 0.341 & 0.288 & 0.373 \\
\hline
CNNIQA \cite{cnnIqa} & 0.259 & 0.242 & 0.266 & 0.223 \\
NIMA \cite{nima} & 0.521 & 0.609 & 0.583 & 0.639 \\
\hline
Baseline Model (Sec.~\ref{sec:baseline}) & 0.525 & 0.599 & 0.571 & 0.623 \\
RoIPool Model (Sec.~\ref{sec:p2p_model}) & 0.541	& 0.618 & 0.576 & 0.655 \\
Feedback Model (Sec.~\ref{subsec:patchAugQuality}) & \textbf{0.562}	& \textbf{0.649} & \textbf{0.601} & \textbf{0.685} \\
\hline
\end{tabular}
\caption{\footnotesize{\textbf{Picture quality predictions: }
Performance of picture quality models on the full-size validation and test pictures in the new database. A higher value indicates superior performance. NIQE is not trained. \vspace{-0.5em}}}
\vspace{-1.5em}
\label{tbl:onFlive}
\end{table}

Table \ref{tbl:patches} shows the performances of the \textit{same} trained, unmodified models on the associated picture patches of three reduced sizes ($40\%$, $30\%$ and $20\%$ of linear image dimensions). The Baseline Model maintained or slightly improved performance across patch sizes, while NIQE continued to lag, despite the greater subject agreement on reduced-size patches (Sec. \ref{subsec:processScores}). The performance of NIMA suffered as the patch sizes decreased. Conversely, BRISQUE and CNNIQA improved as the patch sizes decreased, although they were trained on whole pictures.
\begin{table*}[h]
\captionsetup{font=scriptsize}
\setlength\extrarowheight{1.0pt}
\centering
\footnotesize
\begin{tabular}{P{3.1cm}||P{0.7cm}|P{0.7cm}||P{0.7cm}|P{0.7cm}||P{0.7cm}|P{0.7cm}||P{0.7cm}|P{0.7cm}||P{0.7cm}|P{0.7cm}||P{0.7cm}|P{0.7cm}}
\hline
& \multicolumn{4}{c||}{(a)} & \multicolumn{4}{c||}{(b)} & \multicolumn{4}{c}{(c)} \\
\hline 
& \multicolumn{2}{c||}{Validation} & \multicolumn{2}{c||}{Test} & \multicolumn{2}{c||}{Validation} & \multicolumn{2}{c||}{Test} & \multicolumn{2}{c||}{Validation} & \multicolumn{2}{c}{Test}\\
\hline
\textbf{Model} & \textbf{SRCC} & \textbf{LCC} & \textbf{SRCC} & \textbf{LCC} & \textbf{SRCC} & \textbf{LCC} & \textbf{SRCC} & \textbf{LCC} & \textbf{SRCC} & \textbf{LCC} & \textbf{SRCC} & \textbf{LCC}\\
\hline
NIQE \cite{niqe} & 0.109 & 0.106 & 0.251 & 0.271 & 0.029 & 0.011 & 0.217 & 0.109 & 0.052 & 0.027 & 0.154 & 0.031\\
BRISQUE \cite{mittal2012no} & 0.384 & 0.467 & 0.433 & 0.498 & 0.442 & 0.503 & 0.524 & 0.556 & 0.495 & 0.494 & 0.532 & 0.526\\
\hline
CNNIQA~\cite{cnnIqa} & 0.438 & 0.400 & 0.445 & 0.373 & 0.522 & 0.449 & 0.562 & 0.440 & 0.580 & 0.481 & 0.592 & 0.475\\
NIMA~\cite{nima} & 0.587 & 0.637 & 0.688 & 0.691 & 0.547 & 0.560 & 0.681 & 0.670 & 0.395 & 0.411 & 0.526 & 0.524\\
\hline
Baseline Model (Sec.~\ref{sec:baseline}) & 0.561 & 0.617 & 0.662 & 0.701 & 0.577 & 0.603 & 0.685 & 0.704 & 0.563 & 0.541 & 0.633 & 0.630\\
RoIPool Model (Sec.~\ref{sec:p2p_model}) & 0.641	& 0.731 & 0.724	& 0.782 & 0.686	& 0.752 & 0.759	& 0.808 & 0.733	& 0.760 & 0.769	& 0.792\\
Feedback Model (Sec.~\ref{subsec:patchAugQuality}) & \textbf{0.658}	& \textbf{0.744} & \textbf{0.726}	& \textbf{0.783} & \textbf{0.698} & \textbf{0.762} & \textbf{0.770}	& \textbf{0.819} & \textbf{0.756}	& \textbf{0.783} & \textbf{0.786}	& \textbf{0.808} \\
\hline
\end{tabular}
\caption{\footnotesize{\textbf{Patch quality predictions: }
Results on (a) the largest patches ($40\%$ of linear dimensions), (b) middle-size patches ($30\%$ of linear dimensions) and (c) smallest patches ($20\%$ of linear dimensions) in the validation and test sets. Same protocol as used in Table \ref{tbl:onFlive}. \vspace{-1em}}}
\label{tbl:patches}
\end{table*}

\subsection{RoIPool : a picture $+$ patches model} \label{sec:p2p_model}
Next, we developed a new type of picture quality model that leverages both picture and patch quality information. Our ``RoIPool Model'' is designed in the same spirit as Fast/Faster R-CNN~\cite{fastRCNN, fasterRCNN}, which was originally designed for object detection. As in Fast-RCNN, our model has an \textit{RoIPool} layer which allows the flexibility to aggregate at both patch and picture-sized scales. However, it differs from Fast-RCNN~\cite{fastRCNN} in three important ways. First, instead of regressing for detecting bounding boxes, we predict full-picture and patch quality. Second, Fast-RCNN performs multi-task learning with two separate heads, one for image classification and another for detection. Our model instead shares a single head between patches and images. This was done to allow sharing of the ``quality-aware” weights between pictures and patches. Third, while both heads of Fast-RCNN operate solely on features from ROI-pooled region proposals, our model pools over the entire picture to conduct global picture quality prediction.

\noindent\textbf{Implementation details:} As in~Sec.~\ref{sec:baseline}, we added an ROIPool layer followed by two fully-connected layers to the pre-trained CNN backbone of ResNet-18. The output size of the RoIPool unit was fixed at $2\times2$. All of the hyper-parameters are the same as detailed in Sec.~\ref{sec:baseline}.

\noindent\textbf{Train and test setup:} Recall that we sampled $3$ patches per image and obtained picture and patch subjective scores (Sec.~\ref{sec:dataset}). During training, the model receives the following input: (a) image, (b) location coordinates \texttt{(left, top, right, bottom)} of all $3$ patches and, (c) ground truth quality scores of the image and patches. At test time, the RoIPool Model can process both pictures and patches of any size. Thus, it offers the advantage of predicting the qualities of patches of any number and specified locations, in parallel with the picture predictions.

\begin{figure}[t]
  \centering
  \vspace{-0.2in}
  \begin{minipage}[b]{.42\linewidth}
    \subcaptionbox{}
      {\hspace{0.1em}\includegraphics[width=0.48\linewidth, trim={2.1in 3.5in 2.55in 2.1in}, clip]{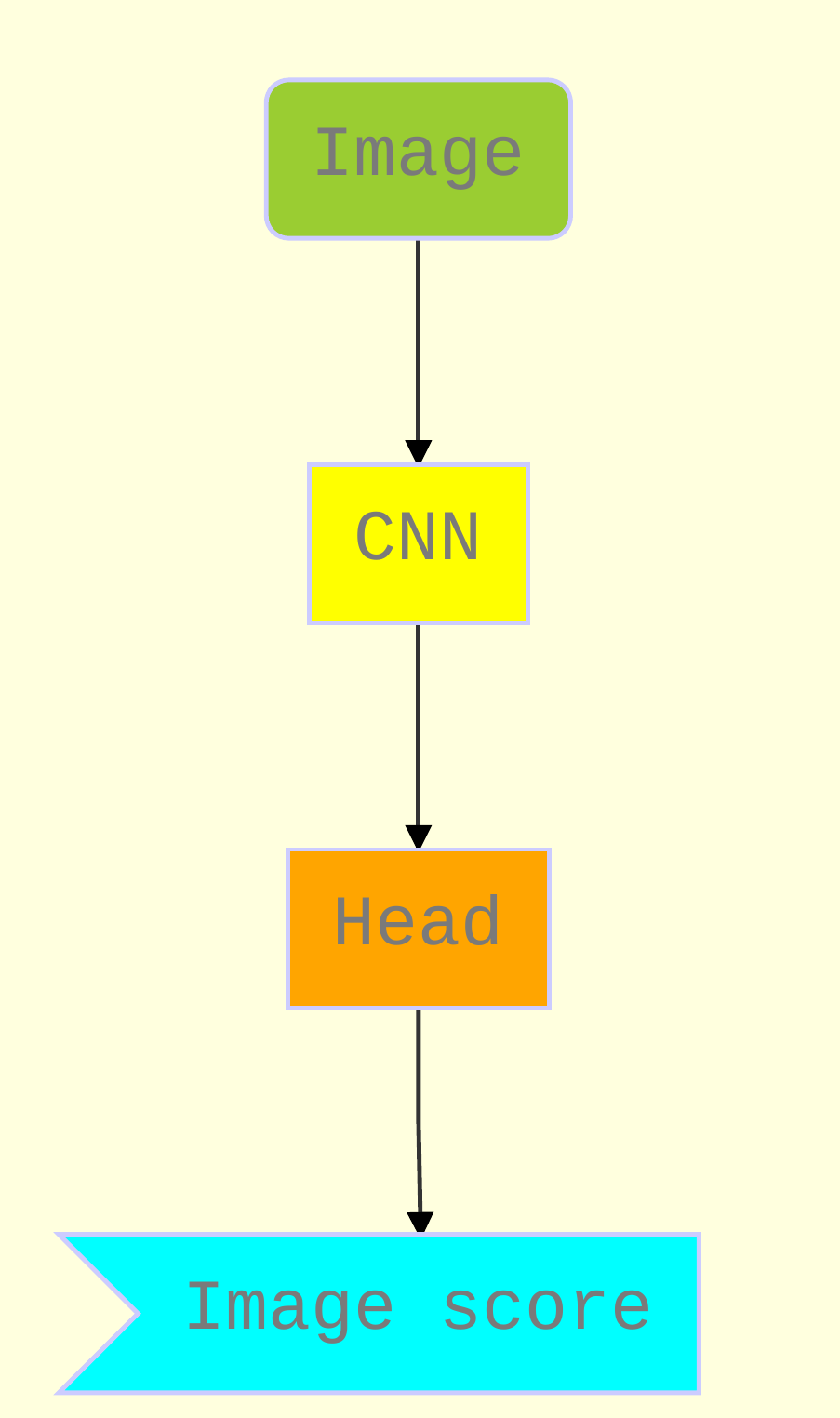}\vspace{-0.5em}}
    \subcaptionbox{}
      {\hspace{-1em}\includegraphics[width=0.75\linewidth, trim={1.8in 2.85in 2.1in 1.8in}, clip]{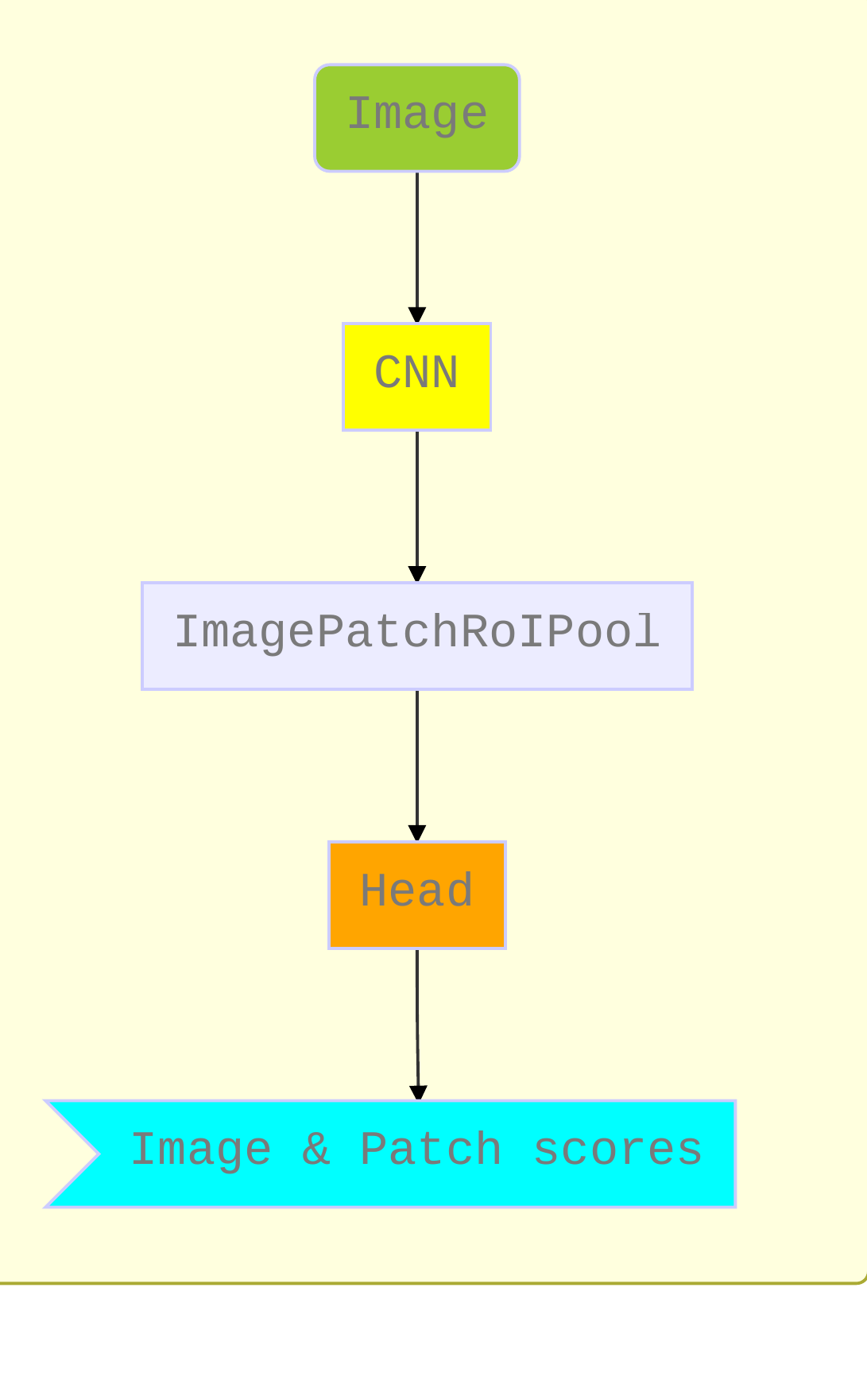}\vspace{-0.5em}}%
  \end{minipage}%
  \hspace{-3.5em}
  \begin{minipage}[t]{.71\linewidth}
    \subcaptionbox{}
      {\includegraphics[width=\linewidth, trim={1.5in 2.8in 1.5in 1.5in}, clip]{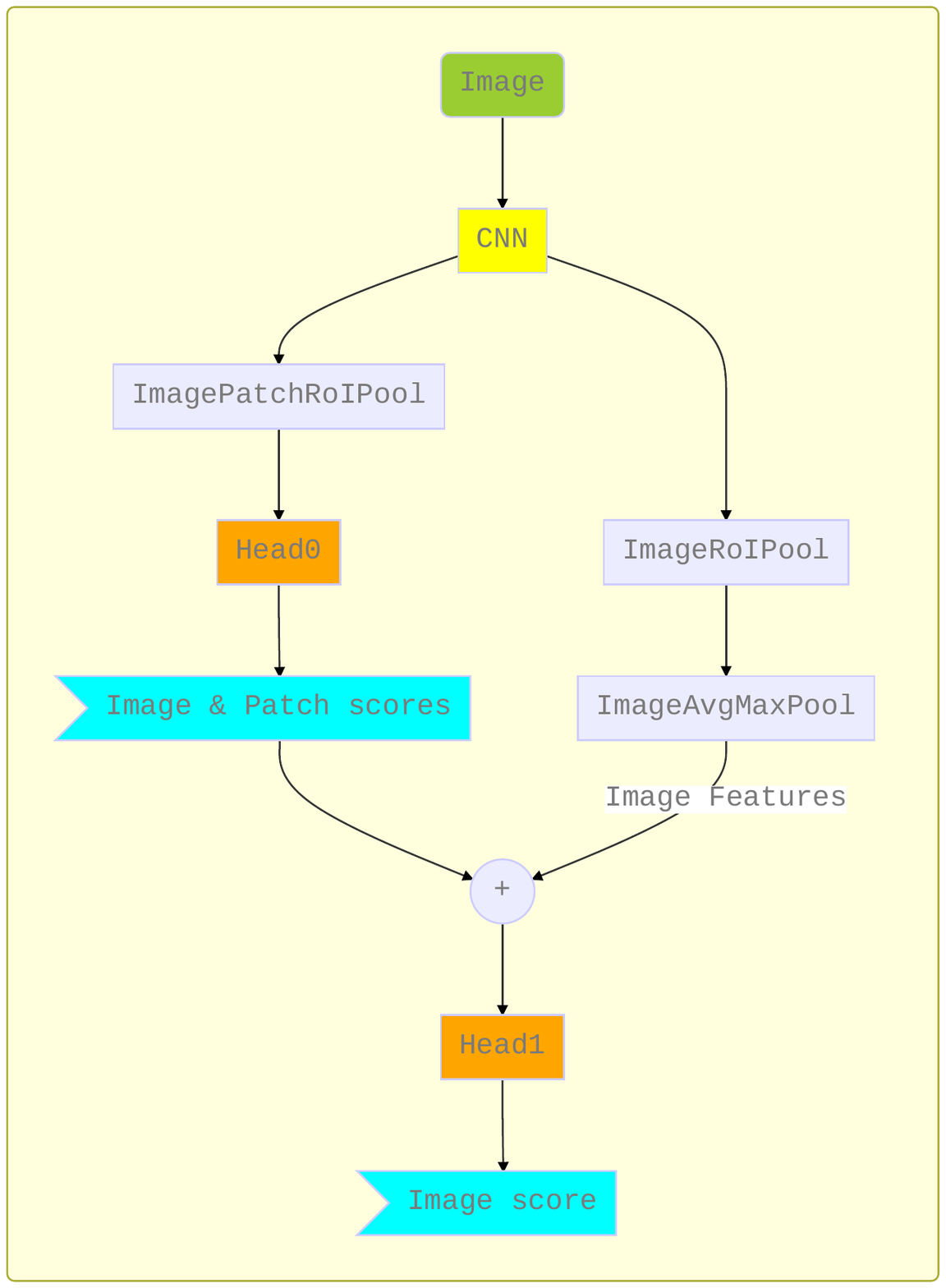}\vspace{-1em}}%
  \end{minipage}%
  \vspace{-0.05in}
  \caption
    {%
      \footnotesize{Illustrating the different deep quality prediction models we studied. (a) \textbf{Baseline Model:} ResNet-$18$ with a modified head trained on pictures (Sec.~\ref{sec:baseline}). (b) \textbf{RoIPool Model:} trained on both picture and patch qualities (Sec.~\ref{sec:p2p_model}). (c) \textbf{Feedback Model:} where the local quality predictions are fed back to improve global quality predictions (Sec.~\ref{subsec:patchAugQuality}).\vspace{-2em}}
      \label{fig:roiPool}%
    }%
\end{figure}

\noindent\textbf{Results:} As shown in Table \ref{tbl:onFlive}, the RoIPool Model yields better results than the Baseline Model and NIMA on whole pictures on both validation and test datasets. When the same trained RoIPool Model was evaluated on patches, the performance improvement was more significant. Unlike the Baseline Model, the performance of the ROIPool model increased as the patch sizes were reduced. This suggests that: (i) the RoIPool Model is more scalable than the Baseline Model, hence better able to predict the qualities of pictures of varying sizes, (ii) accurate patch predictions can help guide global picture prediction, as we show in Sec. \ref{subsec:patchAugQuality}, (iii) this novel picture quality prediction architecture allows computing local quality maps, which we explore next.
 \begin{figure}[h]
\vspace{-1em}
\begin{center}$
\vspace{-0.3em}
\begin{array}{cc}
\includegraphics[width=0.49\linewidth]{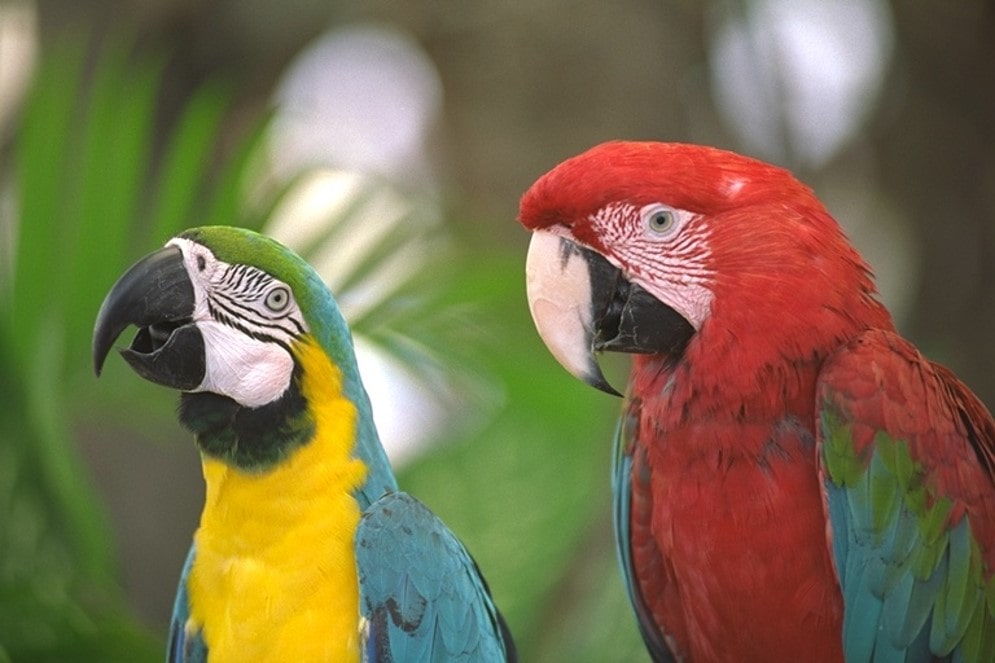} & \hspace{-0.8em} 
\includegraphics[width=0.49\linewidth]{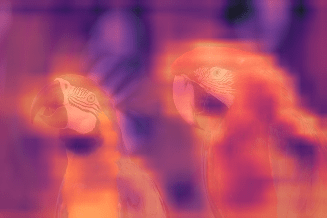} \\
\end{array}$ 
\vspace{-0.3em}
$\begin{array}{cc}
\includegraphics[width=0.49\linewidth]{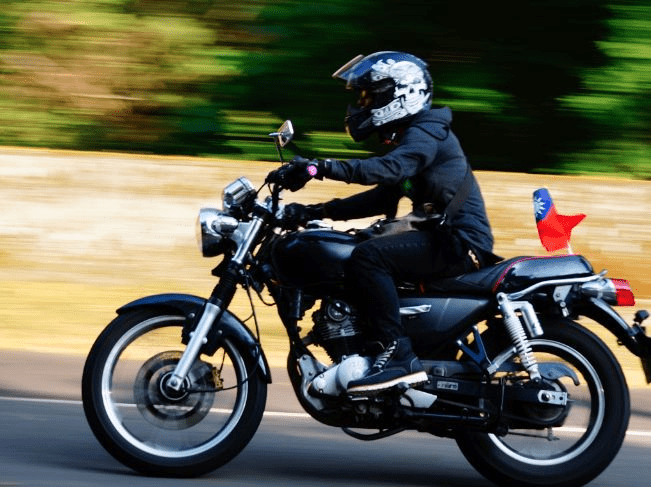} & \hspace{-0.8em} 
\includegraphics[width=0.49\linewidth]{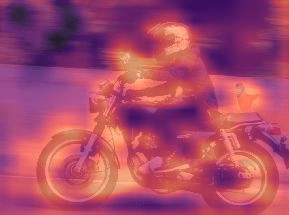} \\
\end{array}$ 
\vspace{-0.3em}
$\begin{array}{cc}
\includegraphics[width=0.49\linewidth]{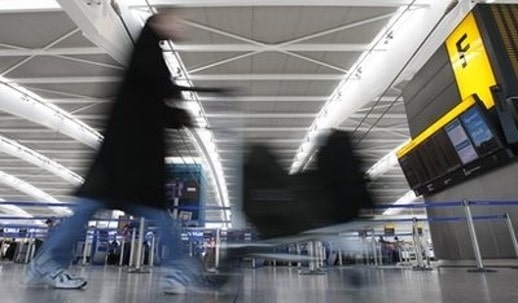} & \hspace{-0.8em} 
\includegraphics[width=0.49\linewidth]{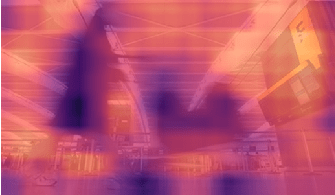} \\
\end{array}$ 
\includegraphics[width=1\linewidth]{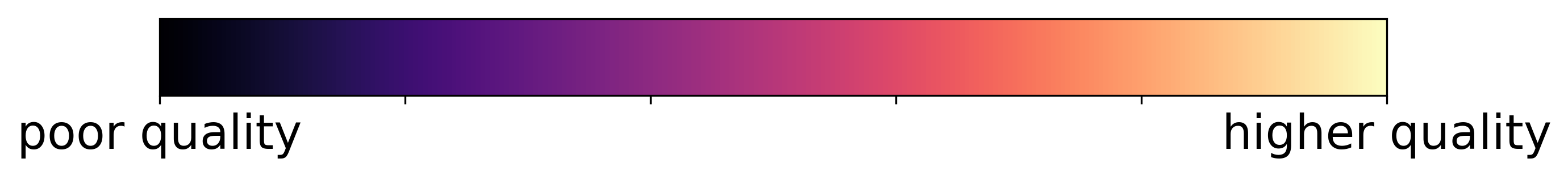} 
\caption{\footnotesize{\textbf{Spatial quality maps} generated using the RoIPool Model (Sec.~\ref{sec:p2p_model}). Left: Original Images. Right: Quality maps blended with the originals using magma color.}}
\vspace{-0.1in}
\label{fig:qualMaps}
\end{center}
\end{figure}
\subsection{Predicting perceptual quality maps} \label{sec:qualityMaps}
Next, we used the ROIPool model to produce patch-wise quality maps on each image, since it is flexible enough to make predictions on any specified number of patches. This unique picture quality map predictor is the first deep model that is learned from true human-generated picture and patch labels, rather than from proxy labels delivered by an algorithm, as in \cite{fullyDeepIQA}. We generated picture quality maps in the following manner: (a) we partitioned each picture into a grid of $32\times32$ non-overlapping blocks, thus preserving aspect ratio (this step can be easily extended to process denser, overlapping, or smaller blocks) (b) Each block's boundary coordinates \texttt{(left, top, right, bottom}) were provided as input to the RoIPool to guide learning of patch quality scores (c) For visualization, we applied bi-linear interpolation to the block predictions, and represented the results as magma color maps. We then $\alpha$-blended the quality maps with the original pictures ($\alpha = 0.8$). 
From Fig.~\ref{fig:qualMaps}, we may observe that the ROIPool Model is able to accurately distinguish regions that are blurred, washed-out, or poorly exposed, from high-quality regions. Such spatially localized quality maps have great potential to support applications like image compression, image retargeting, and so on.


\subsection{A local-to-global feedback model}\label{subsec:patchAugQuality}
As noted in Sec.~\ref{sec:qualityMaps}, local patch quality has a significant influence on global picture quality. Given this, how do we effectively leverage local quality predictions to further improve global picture quality? To address this question, we developed a novel architecture referred to as the Feedback Model (Fig.~\ref{fig:roiPool}(c)). In this framework, the pre-trained backbone has two branches: (i) an RoIPool layer followed by an FC-layer for local patch and image quality prediction (\texttt{Head0}) and (ii) a global image pooling layer. The predictions from \texttt{Head0} are concatenated with the pooled image features from the second branch and fed to a new FC layer (\texttt{Head1}), which makes whole-picture predictions. 

From Tables \ref{tbl:onFlive} and \ref{tbl:patches}, we observe that the performance of the Feedback Model on both pictures and patches is improved even further by the unique local-to-global feedback architecture. This model consistently outperformed \underline{all} shallow and deep quality models. The largest improvement is made on the whole-picture predictions, which was the main goal. The improvement afforded by the Feedback Model is understandable from a perceptual perspective, since, while quality perception by a human is a low-level task involving low-level processes, it also involves a viewer casting their foveal gaze at discrete localized patches of the picture being viewed. The overall picture quality is likely an integrated combination of quality information gathered around each fixation point, similar to the Feedback Model.

\begin{figure}[t]
\vspace{-1em}
\begin{center}$
\vspace{-0.3em}
\begin{array}{cc}
\hspace{-1em} 
\includegraphics[height=0.39\linewidth, trim = {0.2in, 0.1in, 0.2in, 0.11in}, clip]{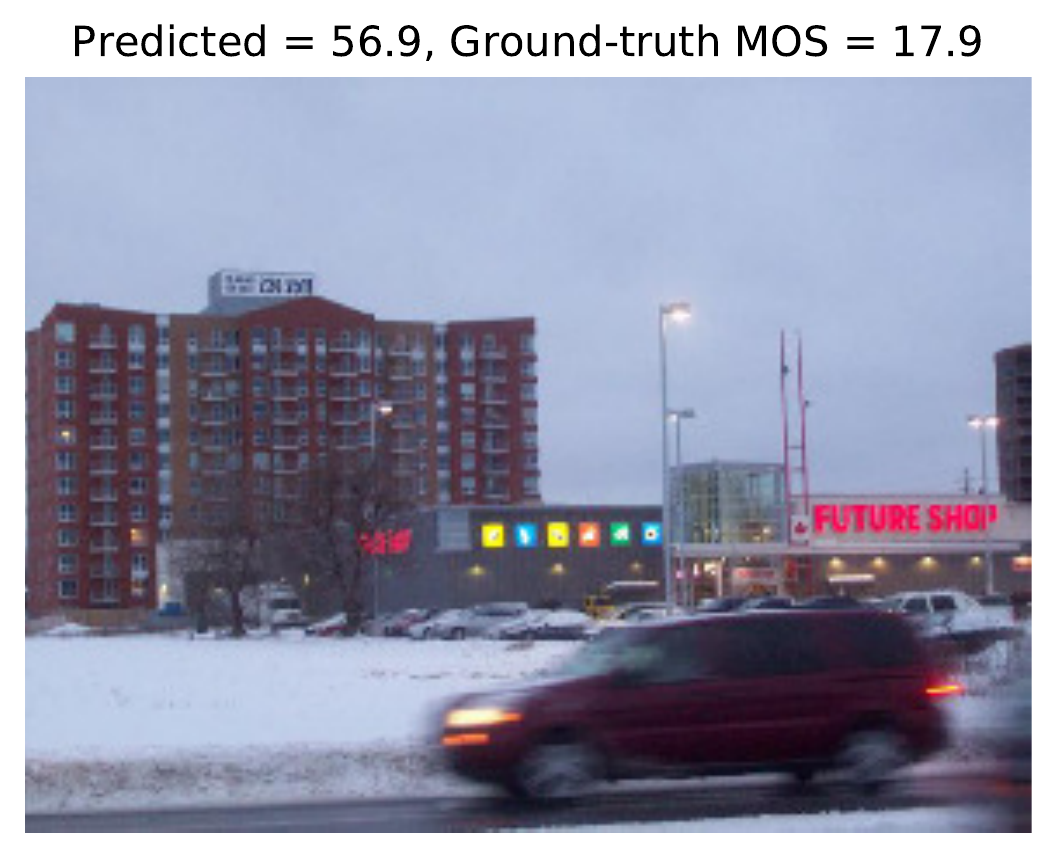} & \hspace{-0.8em} 
\includegraphics[height=0.39\linewidth, trim = {0.1in, 0.1in, 0.2in, 0.11in}, clip]{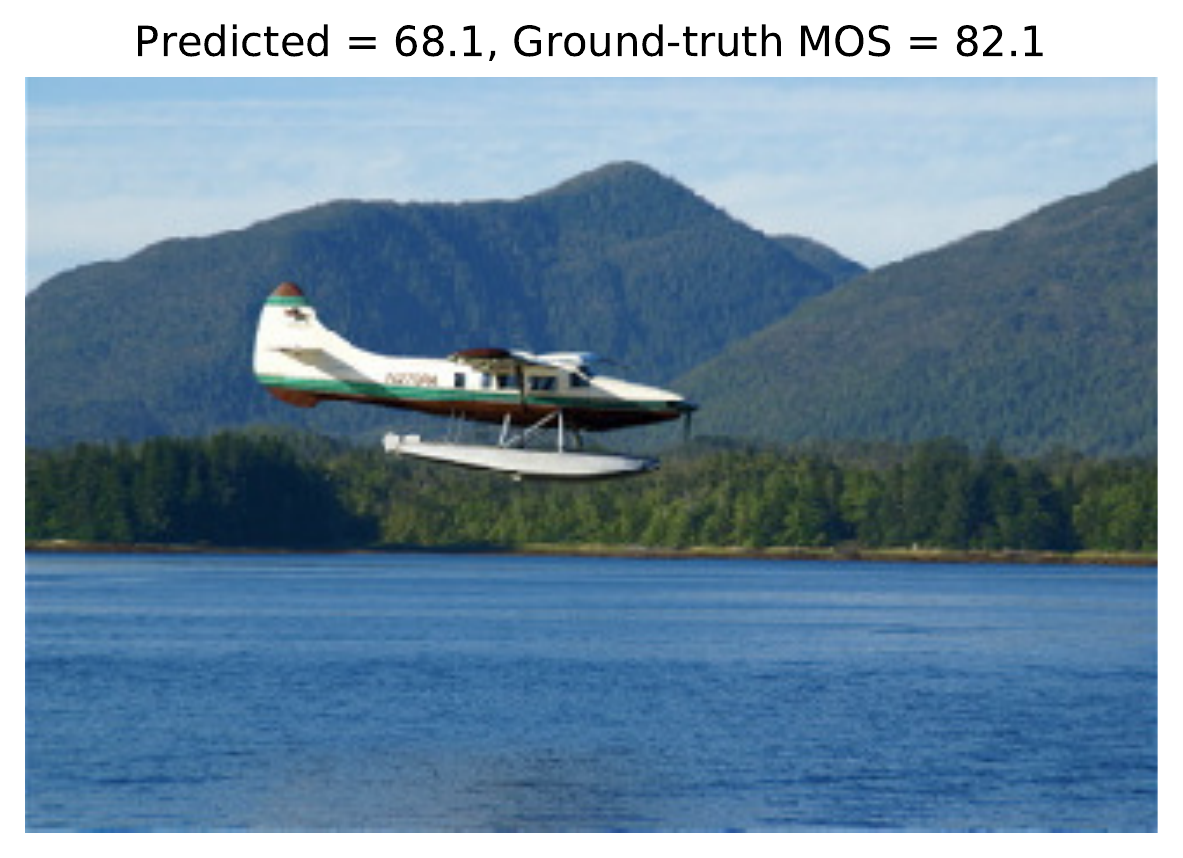} \\
\tiny{(a)} & \tiny{(b)}\\
\vspace{-1.5em}
\end{array}$ 
\caption{\footnotesize{\textbf{Failure cases:} Examples where the Feedback Model's predictions differed the most from the ground truth predictions.}}
\vspace{-0.3in}
\label{fig:failure_eg}
\end{center}
\end{figure}

\noindent \textbf{Failure cases:} While our model attains good performance on the new database, it does make errors in prediction. Fig~\ref{fig:failure_eg}(a) shows a picture that was considered of a very poor quality by the human raters (MOS=$18$), while the Feedback model predicted an overrated score of $57$, which is moderate. This may have been because the subjects were less forgiving of the blurred moving object, which may have drawn their attention. Conversely, Fig~\ref{fig:failure_eg}(b) is a picture that was underrated by our model, receiving a predicted score of $68$ against the subject rating of $82$. It may have been that the subjects discounted the haze in the background in favor of the clearly visible waterplane. These cases further reinforce the difficulty of perceptual picture quality prediction and highlight the strength of our new dataset.

\subsection{Cross-database comparisons} \label{sec:cross_data}
Finally, we evaluated the Baseline (Sec.~\ref{sec:baseline}), RoIPool (Sec.~\ref{sec:p2p_model}), Feedback (Sec.~\ref{subsec:patchAugQuality}) , and other baselines  -- all trained on the proposed dataset -- on two other smaller ``in-the-wild” databases CLIVE \cite{clive} and KonIQ-10k \cite{koniq} \underline{without any fine-tuning}. From Table \ref{tbl:cliveKoniq}, we may observe that all our three models, trained on the proposed dataset, transfer well to other databases. The Baseline, RoIPool, and Feedback Models all outperformed the shallow and other deep models \cite{nima, cnnIqa} on both datasets. This is a powerful result that highlights the representativeness of our new dataset and the efficacy of our models.
The best reported numbers on both databases~\cite{mixedDBiqa} uses a Siamese ResNet-34 backbone by training and testing on the same datasets (along with 5 other datasets). While this model reportedly attains $0.851$ SRCC on CLIVE and $0.894$ on KonIQ-$10$K, we achieved the above results by directly applying pre-trained models, thereby not allowing them to adapt to the distortions of the test data. When we also trained and tested on these datasets, our picture-based Baseline Model also performed at a similar level, obtaining an SRCC of $0.844$ on CLIVE  and $0.890$ on KonIQ-$10$K.
\begin{table}[t]
\captionsetup{font=scriptsize}
\setlength\extrarowheight{1.0pt}
\centering
\footnotesize
\begin{tabular}{P{3.1cm}||P{0.7cm}|P{0.7cm}||P{0.7cm}|P{0.7cm}}
\hline
& \multicolumn{4}{c}{\textbf{Validation Set}} \\
\hline
& \multicolumn{2}{c||}{\textbf{CLIVE} \cite{clive}} & \multicolumn{2}{c}{\textbf{KonIQ} \cite{koniq}} \\
\hline
\textbf{Model} & \textbf{SRCC} & \textbf{LCC} & \textbf{SRCC} & \textbf{LCC} \\
\hline
NIQE \cite{niqe} & 0.503 & 0.528 & 0.534 & 0.509 \\
BRISQUE \cite{mittal2012no} & 0.660 & 0.621 & 0.641 & 0.596 \\
\hline
CNNIQA \cite{cnnIqa} & 0.559 & 0.459 & 0.596	& 0.403 \\
NIMA \cite{nima} & 0.712 & 0.705 & 0.666	& 0.721 \\
\hline
Baseline Model  (Sec.~\ref{sec:baseline}) & 0.740 & 0.725 & 0.753	& 0.764 \\
RoIPool Model (Sec.~\ref{sec:p2p_model}) & 0.762	& \textbf{0.775} & 0.776	& 0.794 \\
Feedback Model (Sec.~\ref{subsec:patchAugQuality}) & \textbf{0.784}	& 0.754 & \textbf{0.788}	& \textbf{0.808} \\
\hline 
\end{tabular}
\caption{\footnotesize{\textbf{Cross-database comparisons:} Results when models trained on the new database are applied on CLIVE \cite{clive} and KonIQ \cite{koniq} \textbf{without fine-tuning.}\vspace{-1.2em}}}
\vspace{-1.5em}
\label{tbl:cliveKoniq}
\end{table}
\section{Concluding Remarks}\label{sec:conclusion}
Problems involving perceptual picture quality prediction are long-standing and fundamental to perception, optics, image processing, and computational vision. Once viewed as a basic vision science modelling problem to improve on weak Mean Squared Error (MSE) based ways of assessing television systems and cameras, the picture quality problem has evolved into one that demands the large-scale tools of data science and computational vision. Towards this end we have created a database that is not only substantially larger and harder than previous ones, but contains data that enables global-to-local and local-to-global quality inferences. We also developed a model that produces local quality inferences, uses them to compute picture quality maps, and global image quality. We believe that the proposed new dataset and models have the potential to enable quality-based monitoring, ingestion, and control of billions of social-media pictures and videos.

Finally, examples in Fig.~\ref{fig:failure_eg} of competing local vs. global quality percepts highlight the fundamental difficulties of the problem of no-reference perceptual picture quality assessment: its subjective nature, the complicated interactions between content and myriad possible combinations of distortions, and the effects of perceptual phenomena like masking. More complex architectures might mitigate some of these issues. Additionally, mid-level semantic side-information about objects in a picture (e.g., faces, animals, babies) or scenes (e.g., outdoor vs. indoor) may also help capture the role of higher-level processes in picture quality assessment.
{\small
\bibliographystyle{unsrt}
\bibliographystyle{ieee}
\bibliography{egbib}
}
\newpage
\newpage
\onecolumn

\def\thesection{\Alph{section}}
\part*{Supplementary Material -- \\ From Patches to Pictures (PaQ-2-PiQ): \\ Mapping the Perceptual Space of Picture Quality}

\section{Performance Summary}
The performance of NIMA~\cite{nima} reported in the paper used a default MobileNet~\cite{mobileNetV2} backbone. For a fair comparison against the proposed family of models which used ResNet-$18$ backbone, we reported the performance of NIMA (ResNet-$18$) on images (Table~\ref{tbl:onFlive}) and patches (Table~\ref{tbl:patches}) of the new datatbase, and also cross-database performance on CLIVE~\cite{clive} and KonIQ-$10$K~\cite{koniq} (Table~\ref{tbl:cliveKoniq}). Given that the proposed models either compete well or outperform other models in all categories further demonstrates their quality prediction strength across multiple databases containing diverse image distortions.

\begin{table}[h]
\captionsetup{font=scriptsize}
\setlength\extrarowheight{1.0pt}
\centering
\footnotesize
\vspace{0.8em}
\begin{tabular}{P{3.1cm}|P{0.85cm}|P{0.85cm}|P{0.85cm}|P{0.85cm}}
\hline
& \multicolumn{2}{c|}{\textbf{Validation Set}} & \multicolumn{2}{c}{\textbf{Testing Set}} \\
\hline
\textbf{Model} & \textbf{SRCC} & \textbf{LCC} & \textbf{SRCC} & \textbf{LCC} \\
\hline
NIMA(MobileNet v2) \cite{nima} & 0.521 & 0.609 & 0.583 & 0.639 \\
NIMA(ResNet 18) \cite{nima} & 0.503 & 0.577 & 0.580 & 0.611 \\
\hline
Baseline Model (Sec. $4.1$) & 0.525 & 0.599 & 0.571 & 0.623 \\
RoIPool Model (Sec. $4.2$) & 0.541	& 0.618 & 0.576 & 0.655 \\
Feedback Model (Sec. $4.4$) & \textbf{0.562}	& \textbf{0.649} & \textbf{0.601} & \textbf{0.685} \\
\hline
\end{tabular}
\caption{\footnotesize{\textbf{Picture quality predictions: }
Performance of picture quality models on the full-size validation and test pictures in the new database. A higher value indicates superior performance. NIQE is not trained. \vspace{-0.5em}}}
\label{tbl:onFlive}
\end{table}
\begin{table*}[h]
\captionsetup{font=scriptsize}
\setlength\extrarowheight{1.0pt}
\centering
\footnotesize
\begin{tabular}{P{3.1cm}||P{0.7cm}|P{0.7cm}||P{0.7cm}|P{0.7cm}||P{0.7cm}|P{0.7cm}||P{0.7cm}|P{0.7cm}||P{0.7cm}|P{0.7cm}||P{0.7cm}|P{0.7cm}}
\hline
& \multicolumn{4}{c||}{(a)} & \multicolumn{4}{c||}{(b)} & \multicolumn{4}{c}{(c)} \\
\hline 
& \multicolumn{2}{c||}{Validation} & \multicolumn{2}{c||}{Test} & \multicolumn{2}{c||}{Validation} & \multicolumn{2}{c||}{Test} & \multicolumn{2}{c||}{Validation} & \multicolumn{2}{c}{Test}\\
\hline
\textbf{Model} & \textbf{SRCC} & \textbf{LCC} & \textbf{SRCC} & \textbf{LCC} & \textbf{SRCC} & \textbf{LCC} & \textbf{SRCC} & \textbf{LCC} & \textbf{SRCC} & \textbf{LCC} & \textbf{SRCC} & \textbf{LCC}\\
\hline
NIMA(MobileNet v2)~\cite{nima} & 0.587 & 0.637 & 0.688 & 0.691 & 0.547 & 0.560 & 0.681 & 0.670 & 0.395 & 0.411 & 0.526 & 0.524\\
NIMA(ResNet 18)~\cite{nima} & 0.578 & 0.600 & 0.676 & 0.696 & 0.516 & 0.505 & 0.672 & 0.657 & 0.324 & 0.316 & 0.504 & 0.483\\
\hline
Baseline Model (Sec. $4.1$) & 0.561 & 0.617 & 0.662 & 0.701 & 0.577 & 0.603 & 0.685 & 0.704 & 0.563 & 0.541 & 0.633 & 0.630\\
RoIPool Model (Sec. $4.2$) & 0.641	& 0.731 & 0.724	& 0.782 & 0.686	& 0.752 & 0.759	& 0.808 & 0.733	& 0.760 & 0.769	& 0.792\\
Feedback Model (Sec. $4.4$) & \textbf{0.658}	& \textbf{0.744} & \textbf{0.726}	& \textbf{0.783} & \textbf{0.698} & \textbf{0.762} & \textbf{0.770}	& \textbf{0.819} & \textbf{0.756}	& \textbf{0.783} & \textbf{0.786}	& \textbf{0.808} \\
\hline
\end{tabular}
\caption{\footnotesize{\textbf{Patch quality predictions: }
Results on (a) the largest patches ($40\%$ of linear dimensions), (b) middle-size patches ($30\%$ of linear dimensions) and (c) smallest patches ($20\%$ of linear dimensions) in the validation and test sets. Same protocol as used in Table \ref{tbl:onFlive}. \vspace{-1em}}}
\label{tbl:patches}
\end{table*}
\begin{table}[h]
\captionsetup{font=scriptsize}
\setlength\extrarowheight{1.0pt}
\centering
\footnotesize
\begin{tabular}{P{3.1cm}||P{0.7cm}|P{0.7cm}||P{0.7cm}|P{0.7cm}}
\hline
& \multicolumn{4}{c}{\textbf{Validation Set}} \\
\hline
& \multicolumn{2}{c||}{\textbf{CLIVE} \cite{clive}} & \multicolumn{2}{c}{\textbf{KonIQ} \cite{koniq}} \\
\hline
\textbf{Model} & \textbf{SRCC} & \textbf{LCC} & \textbf{SRCC} & \textbf{LCC} \\
\hline
NIMA(MobileNet v2) \cite{nima} & 0.712 & 0.705 & 0.666	& 0.721 \\
NIMA(ResNet 18) \cite{nima} & 0.707 & 0.645 & 0.707	& 0.679 \\
\hline
Baseline Model  (Sec. $4.1$) & 0.740 & 0.725 & 0.753	& 0.764 \\
RoIPool Model (Sec. $4.2$) & 0.762	& \textbf{0.775} & 0.776	& 0.794 \\
Feedback Model (Sec. $4.4$) & \textbf{0.784}	& 0.754 & \textbf{0.788}	& \textbf{0.808} \\
\hline 
\end{tabular}
\caption{\footnotesize{\textbf{Cross-database comparisons:} Results when models trained on the new database are applied on CLIVE \cite{clive} and KonIQ \cite{koniq} \textbf{without fine-tuning.}\vspace{-1.2em}}}
\label{tbl:cliveKoniq}
\end{table}

\section{Information on Model Parameters}

Table~\ref{tbl:params} summarizes the number of learnable parameters used by each of the compared models. 
\begin{packed_enum}
\item CNNIQA's~\cite{cnnIqa} poor performance can be attributed to its shallow CNN-based architecture with less than $1$M parameters indicating its inability to model the complex problem. 
\item It is interesting to note that NIMA (MobileNet-v2) performed consistently at par with NIMA (ResNet-$18$) even though it used only $20\%$ of the total parameters.
\item Although RoIPool Model used the same number of parameters as the Baseline Model, it achieved significantly better performance suggesting the importance of accurate local quality predictions for global quality.  
\end{packed_enum}
\begin{table}[h]
\captionsetup{font=scriptsize}
\footnotesize
\centering
\begin{tabular}{c|c|c|c}
\hline
\textbf{Model}               & \textbf{Backbone params} & \textbf{Head params} & \textbf{Total params} \\ \hline
CNNIQA~\cite{cnnIqa} &         -  & -     & 724.90 K       \\
NIMA (MobileNet v2)~\cite{nima} & 2.22 M          & 10.11 K     & 2.23 M       \\
NIMA (ResNet-$18$)~\cite{nima}    & 11.17 M         & 10.11 K     & 11.18 M      \\
\hline
Baseline (Sec. $4.1$)           & 11.17 M         & 537.99 K    & 11.70 M      \\
RoIPool Model (Sec. $4.2$)      & 11.17 M         & 537.99 K    & 11.70 M      \\
Feedback Model (Sec. $4.4$)     & 11.17 M         & 1.07 M      & 12.24 M      \\ \hline
\end{tabular}
\caption{\footnotesize{{Number of model parameters. }}\vspace{-1em}}
\label{tbl:params}
\end{table}

\section{Picture MOS vs Patch MOS scatter plots}
\begin{figure}[h]
\begin{center}
$\begin{array}{cc}
\includegraphics[height= 0.35\linewidth,width=0.46\linewidth]{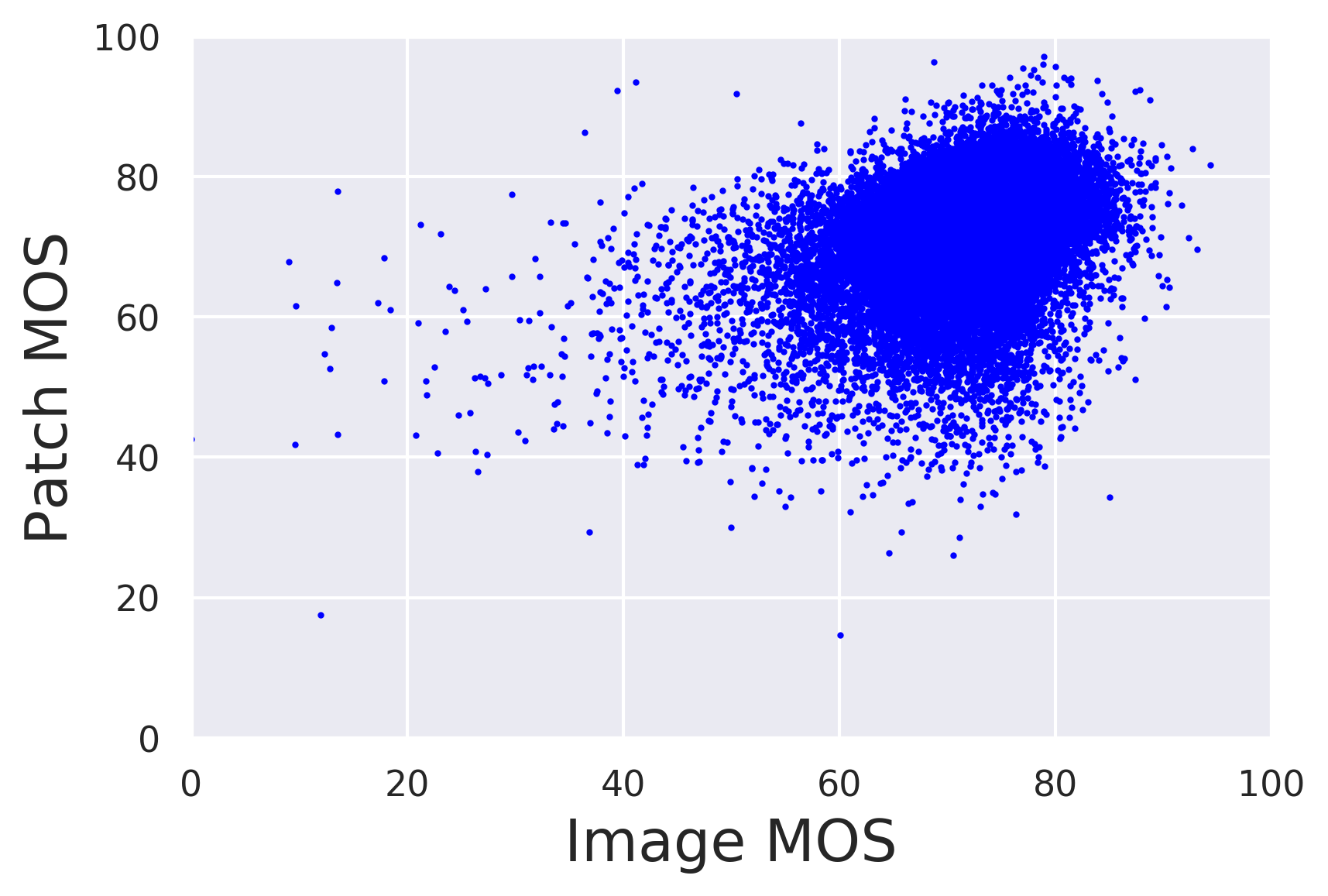} & 
\hspace{-1em}
\includegraphics[height= 0.35\linewidth,width=0.46\linewidth]{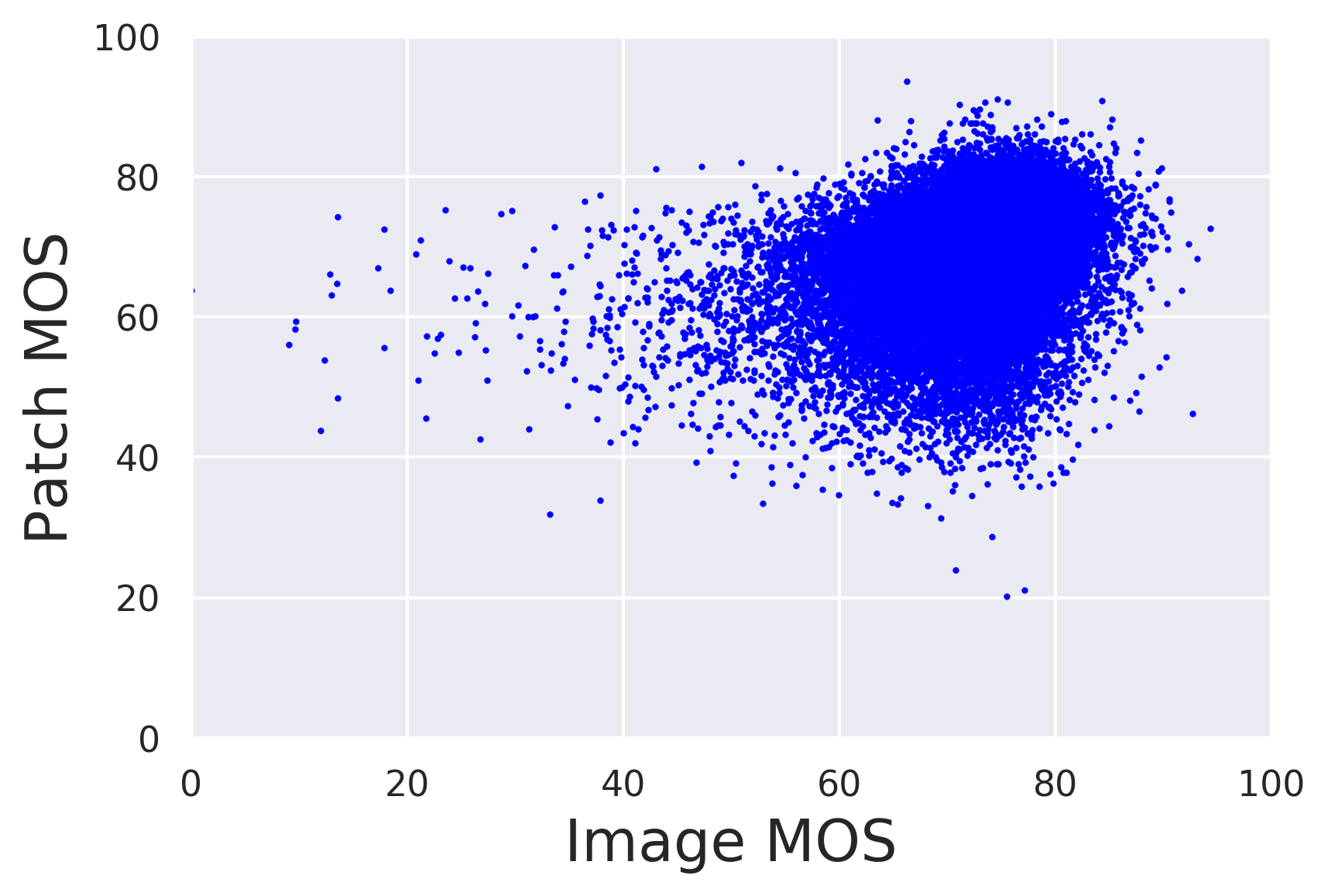} \\
\end{array}$ 
\vspace{-1em}
\caption{\scriptsize{\textbf{Scatter plots of picture MOS vs patch MOS}. Left: Scatter plot of picture MOS vs MOS of second largest patch ($30\%$ of linear dimension) cropped from each same picture.  Right: Scatter plot of picture MOS vs MOS of smallest patch ($20\%$ of linear dimension) cropped from each same picture. 
}}
\vspace{-2em}
\label{fig:patchCorrel}
\end{center}
\end{figure}

\section{Amazon Mechanical Turk Interface}

\begin{figure}[h]
\begin{center}
\includegraphics[width=0.9\linewidth, height=\textheight, trim={0em 0em 1.5em 0em},clip]{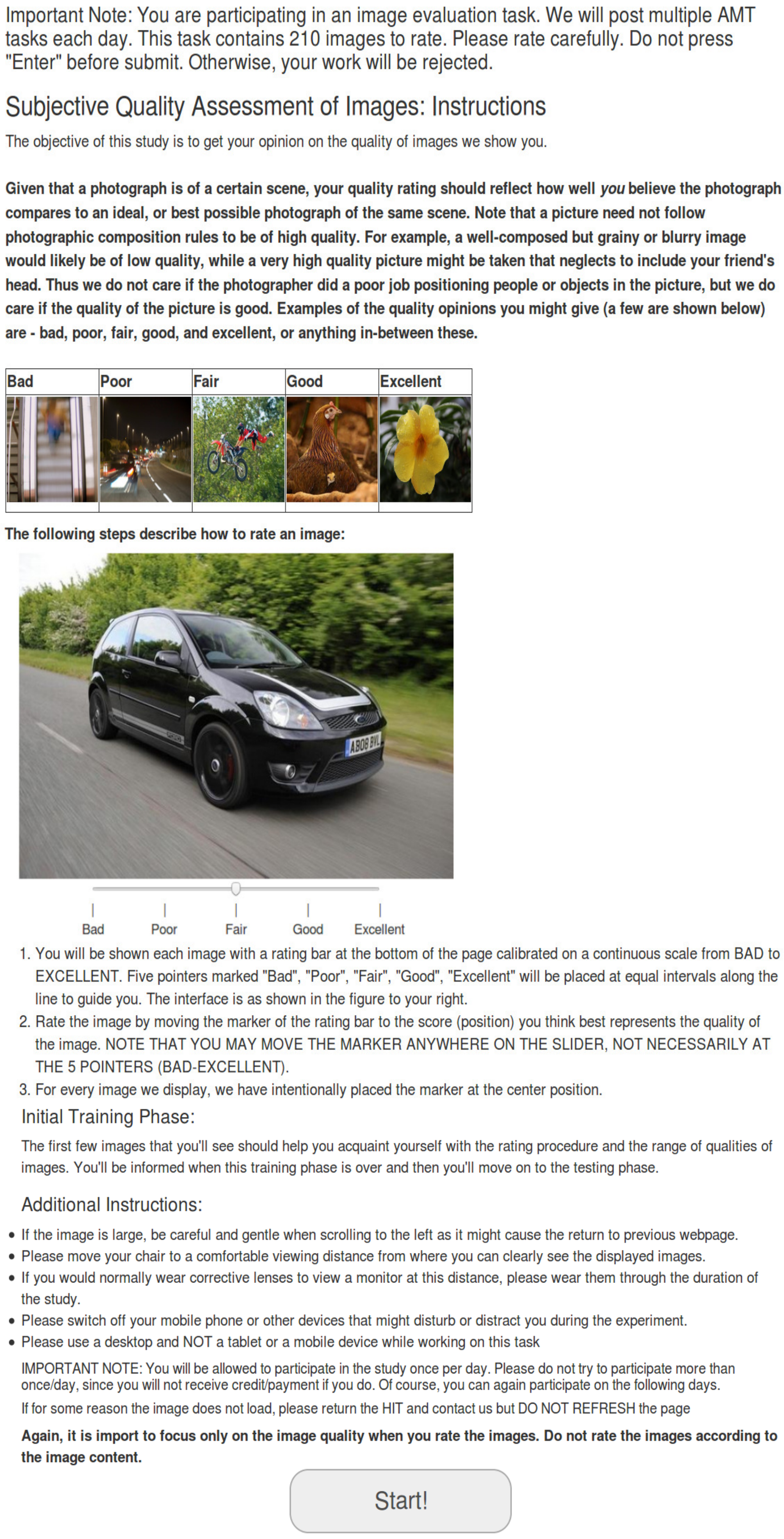}
\caption{\scriptsize{\textbf{AMT task:} The ``Instructions" page shown to workers at the beginning of each HIT.}}
\label{fig:instructions}
\end{center}
\end{figure}

\begin{figure}[h]
\begin{center}
\includegraphics[width=\linewidth, trim={0em 0em 1.5em 0em},clip]{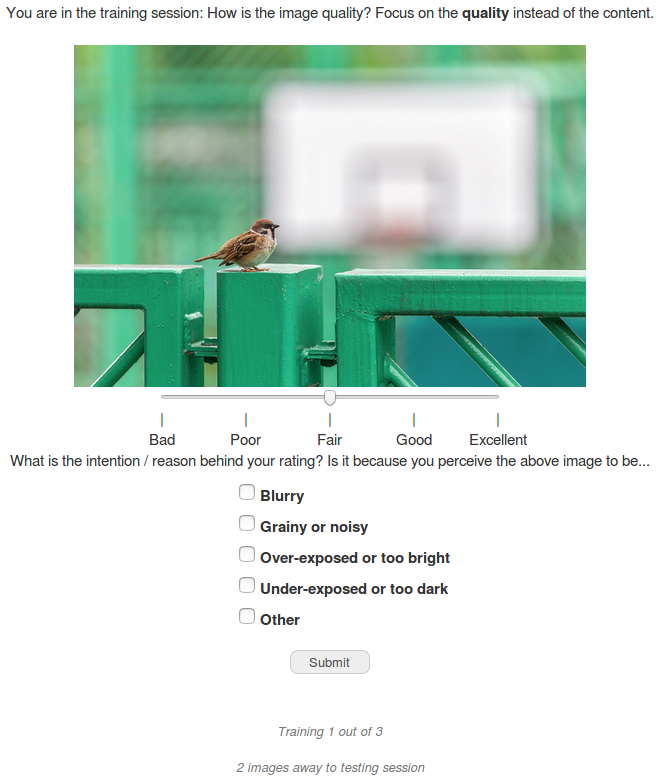}
\vspace{-2em}
\caption{\scriptsize{\textbf{AMT task:} Training session interface of AMT task experienced by crowd-sourced workers when rating pictures.}}
\vspace{-2em}
\label{fig:training1}
\end{center}
\end{figure}

\begin{figure}[h]
\begin{center}
\includegraphics[width=\linewidth, trim={0em 0em 1.5em 0em},clip]{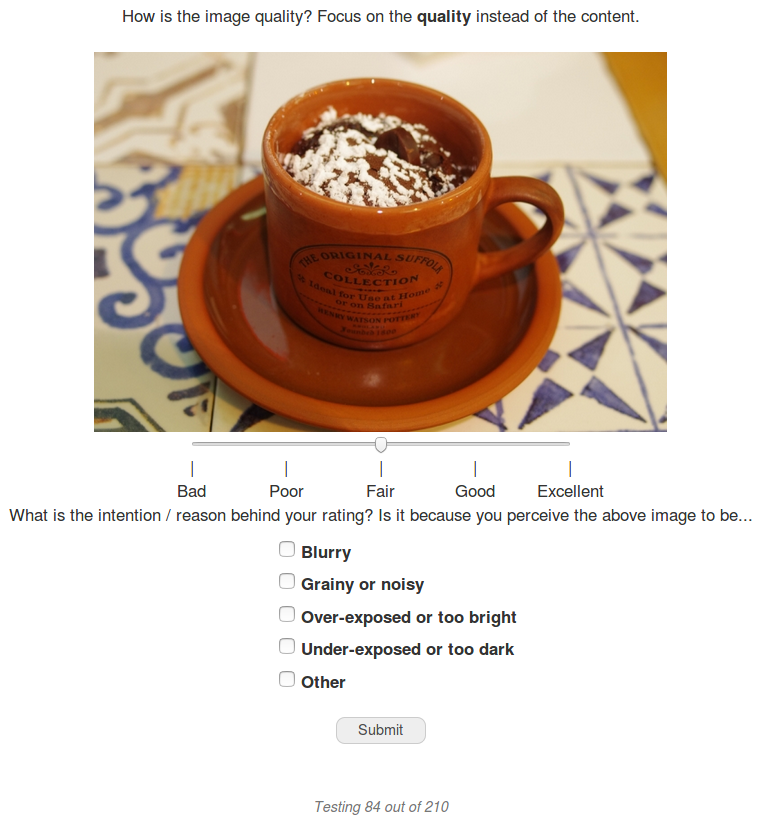}
\vspace{-2em}
\caption{\scriptsize{\textbf{AMT task:} Testing session interface of AMT task experienced by crowd-sourced workers when rating pictures.}}
\vspace{-2em}
\label{fig:testing1}
\end{center}
\end{figure}

We allowed the workers on Amazon Mechanical Turk (AMT) to preview the ``Instructions" page (as shown in Fig~\ref{fig:instructions}) before they accept to participate in the study. Once accepted, they were tasked with rating the quality of images on a Likert scale marked with ``Bad", ``Poor", ``Fair", ``Good" and ``Excellent" as demonstrated in Fig.~\ref{fig:training1} and~\ref{fig:testing1}. A similar user interface was used for patch quality rating task.  

\end{document}